\documentclass[conference]{IEEEtran}
\usepackage{times}
\usepackage{multirow}
\usepackage[numbers]{natbib}
\usepackage{multicol}
\usepackage[bookmarks=true]{hyperref}
\usepackage{graphicx}
\usepackage{color}
\usepackage{array}
\usepackage{algorithm}
\usepackage{algorithmic}
\usepackage{amssymb}
\usepackage{amsmath}
\usepackage[justification=centering]{caption}

\usepackage[normalem]{ulem}
\usepackage{float}

\usepackage[font=small,skip=8pt]{caption}
\newcolumntype{P}[1]{>{\centering\arraybackslash}p{#1}}
\newcommand{\name}[0]{\textit{GenAug}}

\begin{document}

% paper title
% \title{Generative Augmentation for Real-World Robot Learning}
\title{GenAug: Retargeting behaviors to unseen situations via \textbf{Gen}erative \textbf{Aug}mentation}

% You will get a Paper-ID when submitting a pdf file to the conference system
% \author{Author Names Omitted for Anonymous Review. Paper-ID [224]}

\author{Zoey Chen$^1$, Sho Kiami$^1$, Abhishek Gupta*$^1$, Vikash Kumar*$^2$
\smallskip 
\\
$^1$University of Washington ~~~
$^2$ Meta AI
\\
\{qiuyuc, shokiami, abhgupta\}@cs.washington.edu ~~~
vikashplus@meta.com
\\[1em]
\textbf{\Large
\href{https://genaug.github.io}{\color{blue}{genaug.github.io}}}
}

% avoiding spaces at the end of the author lines is not a problem with
% conference papers because we don't use \thanks or \IEEEmembership
\makeatletter
\g@addto@macro\@maketitle{
  \begin{figure}[H]
  \setlength{\linewidth}{\textwidth}
  \setlength{\hsize}{\textwidth}
  % \vspace{-10mm}
  \centering
  \resizebox{0.98\textwidth}{!}{\includegraphics[]{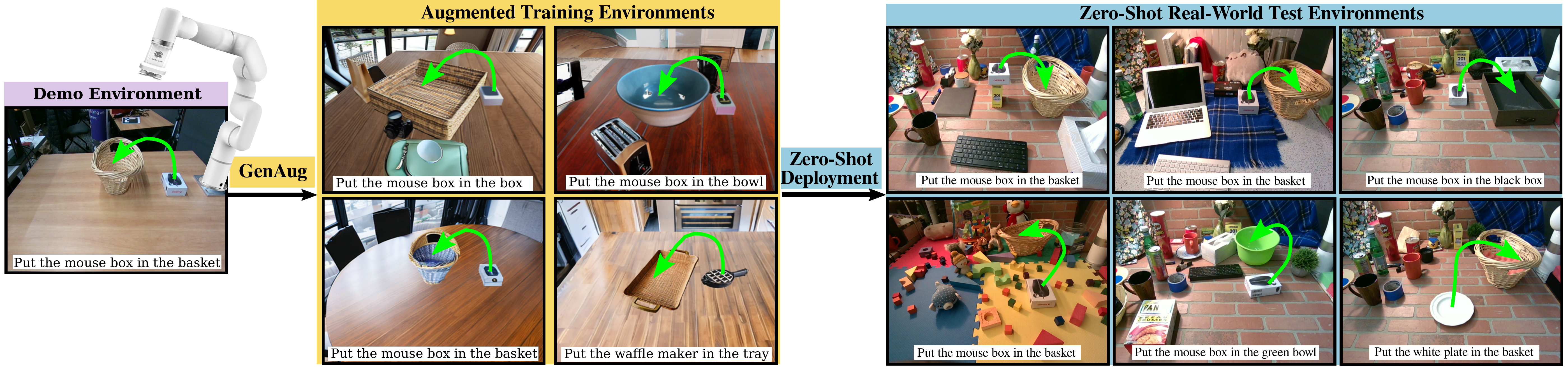}}
  \vspace{-2mm}
  \caption{\footnotesize{An illustration of the problem setting for our proposed system \name{}. \name{} takes a small set of image-action demonstration data on a robotics problem like tabletop pick-and-place and generates a diverse set of augmented image observations to supplement the real-world demonstration dataset. These augmented observations add semantically meaningful visual diversity in objects, distractors, and backgrounds while maintaining functional invariance of the actions. Training on this augmented training set leads to significant improvements in policy generalization, without requiring additional data collection.}}
  \label{fig:main}
  \end{figure}
  \vspace{-4mm}
}
\makeatother

\maketitle
\begin{abstract}
Robot learning methods have the potential for widespread generalization across tasks, environments, and objects. However, these methods require large diverse datasets that are expensive to collect in real-world robotics settings.   
% are severely limited by the amount of data that they are provided or are able to collect. 
% Robots in the real world are likely to only be able to collect a small dataset, both in terms of data quantity and diversity. 
For robot learning to generalize, we must be able to leverage sources of data or priors beyond the robot's own  experience. In this work, we posit that image-text generative models, which are pre-trained on large corpora of web-scraped data, can serve as such a data source. We show that despite these generative models being trained on largely non-robotics data, they can serve as effective ways to impart priors into the process of robot learning in a way that enables widespread generalization. In particular, we show how pre-trained generative models can serve as effective tools for semantically meaningful data augmentation. By leveraging these pre-trained models for generating appropriate ``semantic" data augmentations, we propose a system \name{} that is able to significantly improve policy generalization. We apply \name{} to tabletop manipulation tasks, showing the ability to re-target behavior to novel scenarios, while only requiring marginal amounts of real-world data. We demonstrate the efficacy of this system on a number of object manipulation problems in the real world, showing a 40\% improvement in generalization to novel scenes and objects.

\end{abstract}

\IEEEpeerreviewmaketitle
\section{Introduction}
While robot learning has often focused on the search for optimal behaviors~\cite{levine15endtoend, nagabandi:corl2019} or plans ~\cite{qureshi18mpn}, the power of learning methods in robotics comes from the potential for \emph{generalization}. While techniques such as motion planning or trajectory optimization are effective ways to solve the policy search problem in highly controlled situations such as warehouses or factories, they may fail to generalize to novel scenarios without significant environment modeling and replanning ~\cite{fishman22mpn}. On the other hand, techniques such as imitation learning and reinforcement learning have the \emph{potential} for widespread generalization without significant environment modeling and replanning, especially when combined with deep neural network function approximators ~\cite{kalashnikov18qtopt, brohan23rt1, mahler17dexnet}.
% This potential for generalization makes these techniques important to investigate for robotic deployment into unstructured, real-world environments.
%A natural question is - do the current instantiations of these methods actually live up to the promise of generalization that they make?

Let us consider this question of learning from human demonstrations ~\cite{pomerleau88alvinn}. While imitation learning methods circumvent the challenges of exploration, these methods often impose a heavy burden on data collection by human supervisors. Human demonstrations are often collected by expensive techniques such as on-robot teleoperation or kinesthetic teaching, which limit the amount of real-world data that can be collected. Beyond the sheer quantity of data, the rigidity of most robotics setups makes it non-trivial to collect \emph{diverse} data in a wide variety of scenarios. As a result, many robotics datasets involve a single setup with just a few hours of robot data. This is in stark contrast to the datasets that are common in vision and language problems~\cite{schuhmann22laion, deng09imagenet}, both in terms of quantity \emph{and} the diversity of the data. Given how important large-scale data has been for generalization in these domains, robot learning is likely to benefit from access to a similar scale of data.

Data augmentation can provide a way to improve model generalization, but these techniques typically perform augmentation in low-level visual space, performing operations such as color jitter, Gaussian blurring, and cropping, among others. While this may help with generalization to low-level changes in scene appearance, they are unable to handle large semantic differences in the scene such as distractors, background changes, or object appearance changes. In this work, we aim to provide \emph{semantic} data augmentation to enable broad robot generalization, by leveraging pre-trained generative models. While on-robot data can be limited, the data that pre-trained generative models are exposed to is significantly larger and more diverse ~\cite{schuhmann22laion, deng09imagenet}. Our work aims to leverage these generative models as a source of data augmentation for real-world robot learning. The key idea of our work builds on a simple intuition: demonstrations for solving one task in one environment should still largely be applicable to the same task in new environments despite the visual changes in scenes, background, and object appearances. The small amount of on-robot experience provides demonstrations of the desired behavior, while a generative model can be used to generate widely varying visual scenes, with diverse backgrounds and object appearances under which the same behavior will still be valid. Furthermore, since such generative models are trained on realistic data, the generated scenes are visually realistic and extremely diverse. This allows us to cheaply generate a large quantity of \emph{semantically augmented} data from a small number of demonstrations, providing a learning agent access to significantly more diverse scenes than the purely on-robot demonstration data. As we show empirically, this can lead to widely improved generalization, with minimal additional burden on human data collection. 

We present GenAug, a semantic data augmentation framework that leverages pre-trained text-to-image generative models for real-world robot learning via imitation. Given a dataset of image-action examples provided on a real robot system, GenAug automatically generates ``augmented" RGBD images for entirely different and realistic environments, which display the visual realism and complexity of scenes that a robot might encounter in the real world. In particular, GenAug leverages language prompts with a generative model to change object textures and shapes, adding new distractors and background scenes in a way that is physically consistent with the original scene, for table-top manipulation tasks with a real robot. 
% /ZC{repeat with previous sentence} In comparison to standard techniques for visual data augmentation such as pixel-wise or spatial augmentation, pre-trained generative models ensure semantic and visual realism that a robot might encounter at test time. 
We show that training on this \emph{semantically augmented} dataset significantly improves the generalization capabilities of imitation learning methods on entirely unseen real-world environments, with only 10 real-world demonstrations collected in a single, simple environment. 

In summary, our key contributions are:
\begin{enumerate}
\item We present a general framework for leveraging generative models for data augmentation in robot learning. 
\item We show how this framework can be instantiated in the context of tabletop manipulation tasks in the real world, building on the framework of CLIPort introduced in ~\cite{shridhar2021cliport}. 
\item We demonstrate that GenAug policies can show widespread real-world generalization for tabletop manipulation, even when they are only provided with a few demonstrations in a simple training environment.
\item We provide a number of ablations and visualizations to understand the impact of various design decisions on learned behavior. 
\end{enumerate}

% \begin{figure*}[!h]
%     \centering
%     \includegraphics[width=\textwidth]{figures/system2.png}
%     \caption{GenAug provides options to \VK{reframing an experience in new context by varying objects being interacted, scene being interacted with and the environment in which the scene is present} augmenting texture of the object, change table and room, add distractors and change object category. \VK{Present it in 3 panes (Original demo) -to- (collection of augmentaitons) -to- (retargeted to new scene). FuncAug is about retargetting behaviors to new scenes}}
%     \label{fig:system}
% \end{figure*}
\begin{figure*}[!h]
\vskip -1em 
    \centering
    \includegraphics[width=\textwidth]{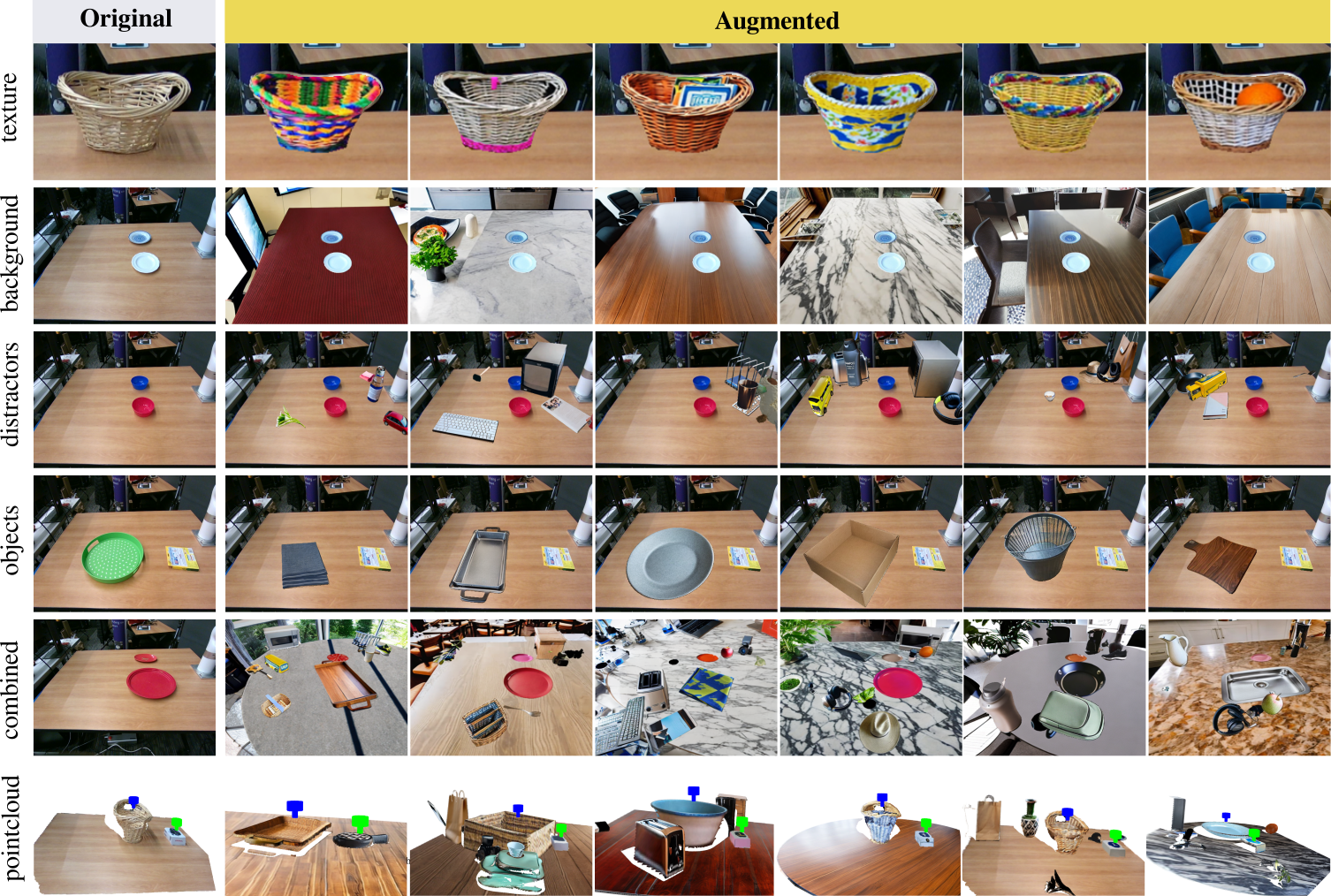}
    \caption{\name{} provides the ability to augment the scene by changing the object texture (first row), changing the background (second row), adding distractors (third row) and changing object categories (fourth row)}
    \vspace{-1em}
    \label{fig:augmented_scene}
\end{figure*}
\section{GenAug: Generative Augmentation for Real-World Data Collection}

\label{sec:method}
In this section, we will describe the problem statement we consider in our semantic data augmentation technique - Generative Augmentation (\name), show how generative models can conceptually be used to inject semantic invariances into robot learning and instantiate a concrete version of this setup for learning policies for tabletop robotic manipulation tasks.

\subsection{Problem Formulation}
In this work, we consider robotic decision-making problems, specifically in robotic manipulation. For the sake of exposition, let us consider prediction problems where an agent is provided access to sensory observations $o \in \mathcal{O}$ (e.g. camera observations) and must predict the most appropriate action $a \in \mathcal{A}$ (e.g. where to move the robot arm for picking up an object). The goal is to learn a predictive model $f_\theta: \mathcal{O} \rightarrow \Delta{A}$ (where $\Delta{A}$ denotes the simplex over actions) that predicts a distribution over actions such that the action $a \sim f_\theta(.|o)$ is able to successfully accomplish a task when executed in the environment. In this work, we will restrict our consideration to supervised learning methods for learning $f_\theta(.|o)$. We will assume that a human expert provides a dataset of optimal data $\mathcal{D} = \{(o_0, a_0), (o_1, a_1), \dots, (o_N, a_N)\}$, and learn policies with standard maximum likelihood techniques \cite{zeng2020transporter, shridhar2021cliport}: 

\begin{equation}
    \max_\theta \mathbb{E}_{(o, a) \sim \mathcal{D}}\left[ \log f_\theta(a|o) \right]
\end{equation}

This training process is limited to the training dataset $\mathcal{D}$ that is actually collected by the human supervisor. Since this might be quite limited, data augmentation methods apply augmentation functions $q: \mathcal{O} \times \mathcal{A} \times \mathcal{Z} \rightarrow \mathcal{O} \times \mathcal{A}$ which generate augmented data $(o', a') = q(o, a, z); z \sim p(z)$, where different noise vectors $z$ generate different augmentations. This could include augmentations like Gaussian noise, cropping, and color jitter amongst others \cite{benton2020learning, cubuk2018autoaugment, shorten2019survey, perez2017effectiveness}. Given an augmentation function, an augmented dataset can be generated $\mathcal{D}_{\text{aug}} = \mathcal{D} \cup \{(o', a')_i\}_{i=1}^M$, where $M \gg N$, and then used for maximum likelihood training of $f_\theta(a|o)$. Typically these augmentation functions $q$ are not learned but instead hand-specified by an algorithm designer, with no real semantic meaning.
% \VK{$g$ for Physics based domain randomization is semantically meaningful.}. 
Instead, they impose invariances to the corresponding disturbances such as color variations, rotations and so on to help combat overfitting. Next, we describe how generative models can be leveraged for semantic data augmentation.

\subsection{Leveraging Generative Models for Data Augmentation}
While data augmentation methods typically hand-define augmentation functions $(o', a') = q(o, a, z); z \sim p(z)$, the generated data $(o', a')$ may not be particularly relevant to the distribution of real-world data that might be encountered during evaluation. In this case, it is not clear if generating a large augmented dataset $\mathcal{D}_{\text{aug}}$ will actually help learned predictors $f$ generalize in real-world settings. The key insight in \name{} is that pretrained generative models are trained on the distribution $p_{\text{real}}(o)$ of real images (including real scenes that a robot might find itself in). This lends them the ability to generate (or modify) the training set observations $o$ in a way that corresponds to the distribution of real-world scenes instead of a heuristic approach such as described in \cite{perez2017effectiveness}. We will use this ability to perform targeted data augmentation for improved generalization of the learned predictor $f_\theta$. 

Let us formalize these generative models as $g: \mathcal{T} \times \mathcal{O} \times \mathcal{Z} \rightarrow \mathcal{O}$, mapping from a text description, an image, and a noise vector to a modified image $o' = g(o, t, z); z \sim p(z)$. This includes commonly used text-to-image inpainting models such as Make-A-Video \cite{singer2022make}, DALL-E 2 \cite{ramesh2022hierarchical}, Stable Diffusion \cite{rombach2022high} and Imagen \cite{saharia2022photorealistic}. It is important to note that since these generative models are simply generating images, they are not able to appropriately generate novel actions $a$, simply novel observations $o$. This suggests that data generated by these generative models will be able to impose \emph{semantic invariance} on the learned model $f_\theta$, i.e ensure that an equivalence group $\{o, g(t_1, o, z_1), g(t_2, o, z_2), \dots, g(t_M, o, z_M)\}$ all map to the same action $a$. To leverage a pretrained text-image generative model for semantic data augmentation, we can simply generate a large set of semantically equivalent observation-action pairs $\{(o, a), (g(t_1, o, z_1), a), (g(t_2, o, z_2), a), \dots, (g(t_M, o, z_M), a)\}$ for every observation $(o,a) \in \mathcal{D}$ using the generative model $g$. Note that the generative models cannot simply generate arbitrary observations $g(t, o, z)$, but only observations that retain semantic equivalence, i.e the ground truth actions for the generated augmentations $\{o, g(t_1, o, z_1), g(t_2, o, z_2), \dots, g(t_M, o, z_M)\}$ are all $a$. We discuss how to actually instantiate this semantic equivalence in the following section. 
% \VK{This is hard to guarantee in practice, therefore can be risky to claim. They can in principle generate garbage images and leads to (garbage, a) pairs. I think we can only claim that given they are trained on real world data. they have high likelihood of generating meaningful augmentations. Now the question is how can we maximize this likelihood}
\name{} allows us to generate a large dataset of \emph{realistic} data augmentations, that ensures robustness to various realistic scenes that may be encountered at test time, while still being able to perform the designated task. Unlike typical data augmentation with the hand-defined shifts described above, the generated augmented observations $\{g(t_1, o, z_1), g(t_2, o, z_2), \dots, g(t_M, o, z_M)\}$ have high likelihood under the distribution of real images $p_{\text{real}}(o)$ that a robot may encounter on deployment. This ensures that the model generalizes to a wide variety of novel scenes, making it significantly more practical to deploy in real world scenarios, since it will be robust to changes in objects, distractors, backgrounds and other characteristics of an environment. It is important to note the limitations of doing this type of augmentation - it will not be able to generate novel actions $a$, but instead generate invariances to realistic observational disturbances $o' = g(t, o, z)$ that are generated by the text-image generative model. If not performed carefully, these augmentations can also possibly invalidate the original action $a$ due to factors such as physical inconsistencies or collisions. Next, we discuss a concrete instantiation of this framework in the context of tabletop robotic manipulation. 

\subsection{Instantiating \name{} for Tabletop Robotic Manipulation}
We scope our discussion of \name{} for this work to tabletop rearrangement tasks with a robot arm. This problem has a non-trivial degree of complexity when performed from purely visual input, especially in very cluttered and visually rich domains, and constitutes a significant body of work in robot learning \cite{zeng2020transporter, shridhar2021cliport}. In particular, we consider tasks where the observation $o$ is a top-down view of the scene, and the action is a spatial action map over the image, indicating where to pick and place objects with a suction-activated gripper. This builds on the transporter networks\cite{zeng2020transporter, shridhar2021cliport} framework for visual imitation learning. In this section, we describe how to instantiate GenAug for semantic data augmentation for these table-top manipulation problems. 

The important question to answer is \textemdash  ~how do we use and prompt the generative model $g$ to generate the appropriate equivalence set of semantically equivalent augmented states for an observation $o$ - $\{o, g(t_1, o, z_1), g(t_2, o, z_2), \dots, g(t_M, o, z_M)\}$, such that the same action $a$ would apply across all of them? We leverage the fact that for a tabletop manipulation task involving picking and placing objects, the same actions are applicable across a wide range of visual appearances including objects being grasped, distractor objects, target receptacles, and backgrounds, as long as the approximate position of the object of interest and the target remain unchanged. 
% \VK{there is an implicit assumption here that agent's trajectory will remain valid. Perhaps it a small point or can gloss over, given pick-place? For that we need to narrow it down to pick-place before this argument}

Given a pick-and-place task on a tabletop, we can perform data augmentations on the visual appearance of 1) the object being grasped or the target receptacle, 2) distractor objects 3) the background or table. We will next describe how we can use the text-to-image depth-guided image generation for generating \name's augmentations and maximize visual diversity while preserving semantic invariances for each of these types of augmentations. 
% --
% [TODO: Refer to a figure to aid describing how this works]
% [TODO: Need to define object to grasp and target receptacle]
% [TODO: Need to properly define semantic invariance]
\begin{figure}[!h]
\vskip -1em 
    \centering
    \includegraphics[width=0.49\textwidth]{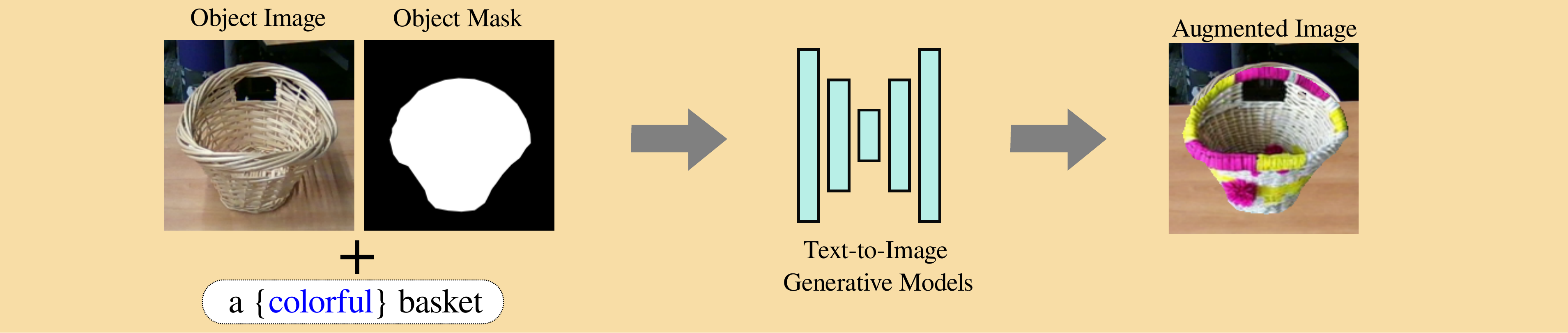}
    \caption{\name{} takes the RGB image and the object mask and uses a depth-guided diffusion model to perform in-category data augmentation.}
    \vspace{-1em}
    \label{fig:system_in_category}
\end{figure}
\subsubsection{Generating Diverse Grasping Objects and Target Receptacles}
\label{sec:system_object}
Simply generating new scenes with an image generation model is unlikely to retain the semantic invariance that we desire for in \name{} since the images will be generated in an uncontrolled way with no regard for functionality. To appropriately retain semantic invariance, we propose a more controlled image generation scheme. In particular, we assume access to masks $\mathcal{M}(o)$ for every observation $o$, labeling the object of interest and the target receptacle. Note that this is only needed for the small number of demonstrations that are collected, not at inference time. To generate a diversity of visuals, we consider augmentation both ``in-category" and ``out-of-category", as described below:
\begin{figure}[h]
% \vskip -0.5em 
    \centering
    \includegraphics[width=0.49\textwidth]{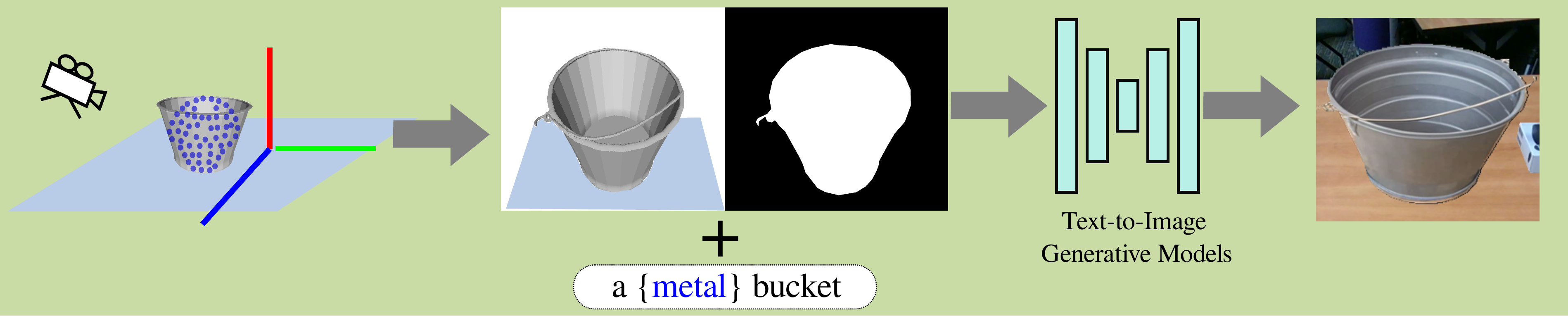}
    \caption{\name{} randomly chooses a new object and place it at the center of the original object to perform cross-category data augmentation}
    \vspace{-1.5em}
    \label{fig:system_cross_category}
\end{figure}

\textbf{In-Category Generation}
For in-category generation, GenAug takes the provided mask $\mathcal{M}(o)$ of the object to grasp (or the target receptacle) and the original RGB image, and applies a pretrained depth-guided text-to-image inpainting model \cite{rombach2021highresolution} to generate novel visual appearances for objects from the same category. To encourage diverse visual appearances of the object, we leverage the fact that the generative model is guided by text and randomly generate novel text prompts involving color (e.g. red, green, yellow, etc) and material (glass, marble, wood, etc) to generate visually diverse objects. Since the object masks remain the same, the resulting positions and shapes are the same, thereby retaining semantic invariance. 
% \VK{model knows the object category? - manually specified?}. 

\textbf{Cross-Category Generation}
While in-category generation provides a degree of visual diversity, it often falls short at generating novel objects altogether and we must consider replacing object categories altogether. When replacing the category of the original object $O_i$ (e.g. a basket) with a new object (e.g. a bucket), one potential technique involves using inpainting models to generate images directly over the masked object (or target receptable). However, naively applying this technique does not generate physically plausible images since it does not appropriately account for geometric consistency during image generation, which causes problems for robotic manipulation. 

To allow the generated images to show physical plausibility and 3-D consistency, we leverage a dataset of object meshes to assist the generative model's generation process. In particular, we first render randomly scaled and sized object meshes from a different category without any visual detail to get the perspective correct using the same camera pose, followed by a process of visual generation with a depth-aware generative model, as described for an in-category generation. The object meshes are able to ensure 3-D consistency and physical plausibility, while the generative model allows for significant visual diversity. We note that since we are doing top-down grasping with a suction gripper, even cross-category generation ensures semantic invariance leaving the point of interaction with the object largely unchanged while boosting visual diversity. 
% \VK{Its not clear to me why this is the case}
\begin{figure}[!h]
\vskip -1em 
    \centering
\includegraphics[width=0.49\textwidth ]{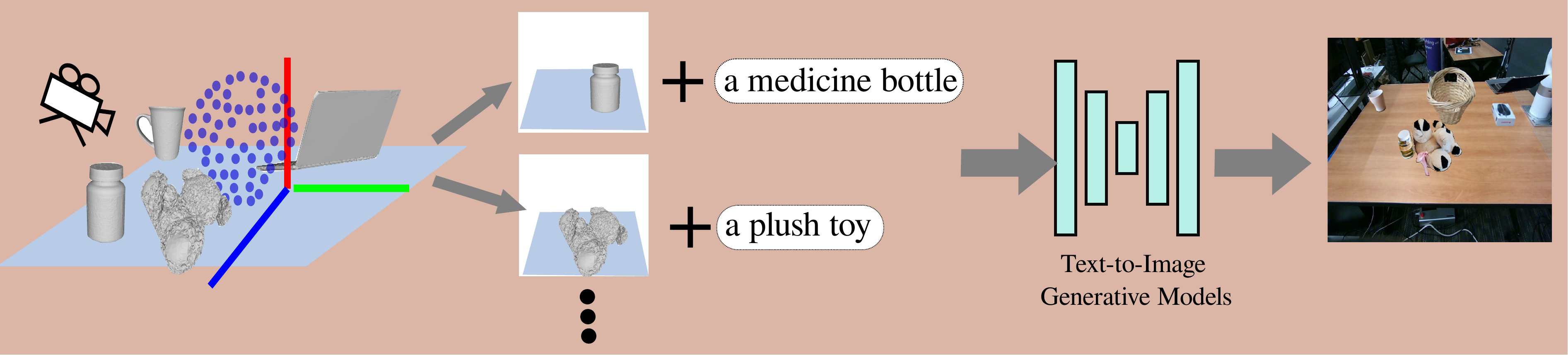}
    \caption{\name{} places a distractor with collision check on the table and  uses a depth-guided model to generate realistic-looking objects that are physically plausible.}
    \vspace{-1em}
    \label{fig:system_distractors}
\end{figure}

\subsubsection{Generating Distractors with Diverse Visual Appearances}
\label{sec:system_distractors}
While Section \ref{sec:system_object} discusses how to augment the appearance of the object to grasp and the target receptacle, real-world scenarios are often cluttered scenes with several irrelevant distractors. GenAug leverages the same techniques described in Section \ref{sec:system_object} to generate scenes with a diversity of visual distractors. Similar to cross-object augmentation, to add a new distractor $D_i$, we randomly choose a new object mesh from a family of object assets and render it on the table, followed by visual generation with a text-to-image generative model as described in Section \ref{sec:system_object}. Importantly since the distractors must be generated in a way that retains semantic invariance, they must not be in collision with the object to grasp or the target receptacle. To ensure this, we compute collisions by checking for overlapping bounding boxes (in image space) between the generated distractor $D_i$ and masks $\mathcal{M}(o)$ for the object to grasp and the target receptacle and remove this distractor if it is in collision. This ensures semantic invariance while being able to generate very cluttered and diverse scenes, as shown in Figure \ref{fig:system_distractors}.  

\begin{figure}[!h]
\vspace{-1em}
    \centering
\includegraphics[width=0.49\textwidth]{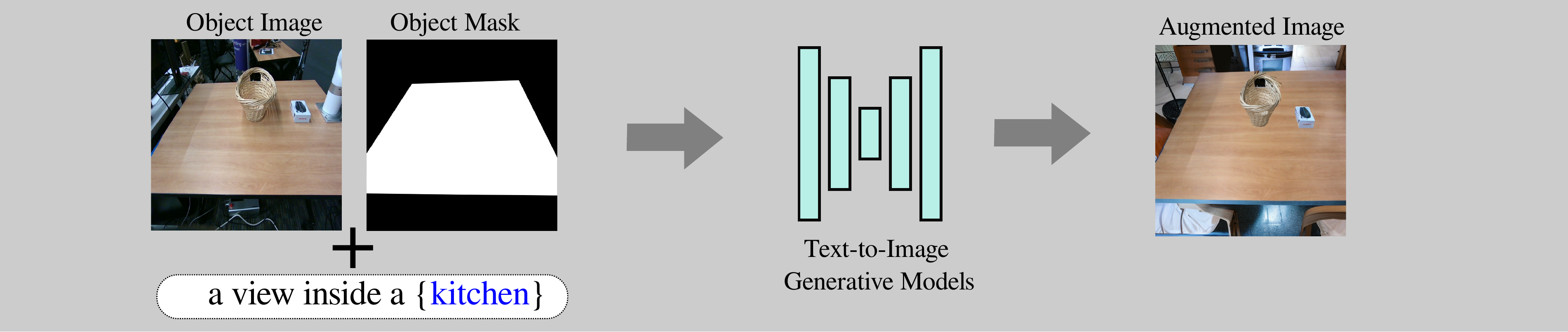}
    \caption{\name{} takes the original RGB image and the mask of the Non-background regions to generate different styles of background scenes such as kitchen, living room, or restaurant.}
    \vspace{-1em}
    \label{fig:system_room}
\end{figure}

\subsubsection{Generating Diverse Backgrounds}
To augment not just the appearances of objects, but also the background of the scene while ensuring semantic invariance, we can simply invert the process of generation in Section \ref{sec:system_object} and \ref{sec:system_distractors}. In particular, we can simply hold the object, target receptacle, and distractor masks fixed while asking a text-guided generative model to generate a diverse range of backgrounds, as shown in Figure \ref{fig:system_room}. Since the object masks are all held fixed, their positions remain invariant, while ensuring that the visual appearance of the background and table varies widely.

GenAug leverages these three forms of semantic augmentation - 1) visual object generation, 2) distractor generation and 3) background generation to augment robot learning data with a large amount of semantically invariant, yet visually diverse data. This data has a significant overlap with the types of environments that might be encountered by a system in the real world, and as we show empirically in Section \ref{sec:experiment}, is able to improve the robustness and generalization of robot learning models significantly. Once data is generated with a combination of these three forms of augmentation, we can then simply run standard maximum likelihood techniques for learning manipulation from the augmented dataset. To enable GenAug to be an effective tool for robot learning, we next describe the setup we used for real-world experiments. 

\section{System Details}
\subsection{Hardware Setup}
\label{sec: system_setup}
\begin{figure}[!h]
\vspace{-1em}
    \centering
    \includegraphics[width=0.48\textwidth]{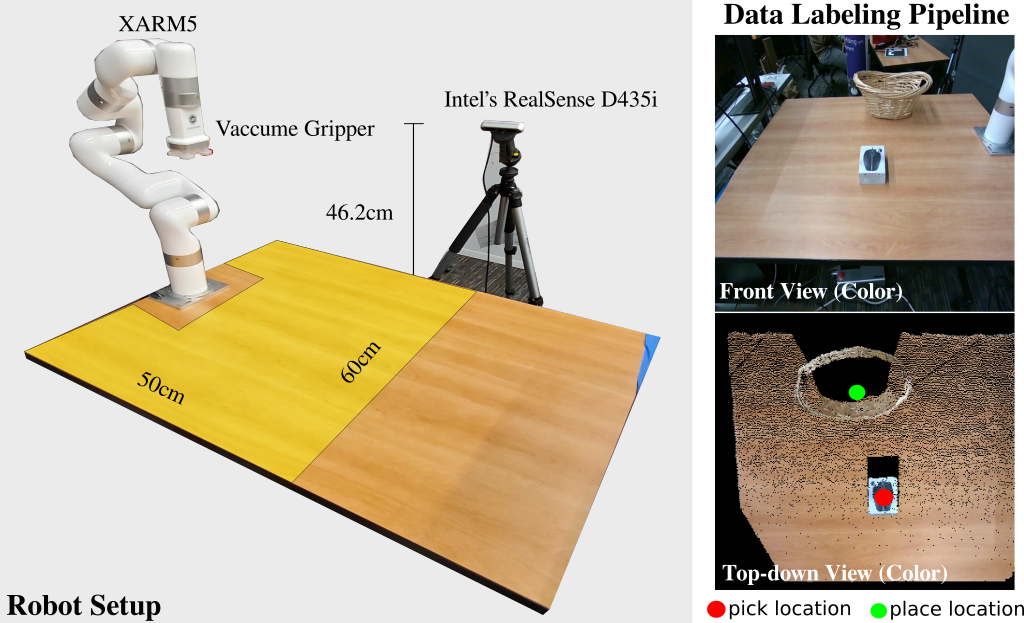}
    \caption{An illustration of our robot experiment setup and data labeling pipeline. A user clicks locations on a top-down view image, to indicate pick (red) and place (green) locations in the robot space.}
    \vspace{-1em}
    \label{fig:setup}
\end{figure}

Since \name{} is instantiated in this work for tabletop manipulation, we use a robot arm equipped with a suction gripper for all our hardware experiments. In particular, we use the 6 DoF xArm5 with a vacuum gripper manipulator, and control it directly in end-effector space. As shown in Figure \ref{fig:setup}, we attached the xArm to the end of a large wooden table in a brightly lit room and set up the depth camera on a tripod on one side of the table so that it has a clear view of the robot and any objects on the table. 

\name{} requires RGBD observations of the demonstration scenes, masks for the object of interest and the target receptacles, as well as a calibrated camera pose. In our real-world setup, we obtain RGBD images from an Intel RealSense Camera (D435i) and manually label the object masks for the collected demonstrations. While the input for \name{} is the camera observations from the RealSense camera, the input observation and the action for the predictive model $f_\theta$ operate on a top-down view, as described in \cite{zeng2020transporter}. 

In order to guide the robot to complete the tasks in the cluttered environment, we largely build on the architecture and training scheme of CLIPort \cite{shridhar2021cliport}, which combines the benefits of language-conditioned policy learning with transporter networks \cite{zeng2020transporter} for data-efficient imitation learning in tabletop settings. \name{} replaces the imitation learning dataset in CLIPort with a much larger augmented one, as described in Section \ref{sec:system_genaug}. Implementation details can be found in Appendix B. 

\subsection{Demonstration Collection}
\label{sec:system_data_collection}

To collect demonstrations, we rely on humans to collect action labels for various pick-and-place tasks. We first project the front camera view to a 2D top-down image and height map of the scene. The user can manually click locations on the top-down images to indicate the pick and place locations. These locations are then converted to end-effector positions in full Cartesian space, which is then provided to a low-level controller that uses inverse kinematics and position control. We save the demonstration if the robot can successfully complete the task. We collect 10 demonstrations per task and 10 tasks in total, all in one single environment as shown as "Demo Environment" in Figure \ref{fig:test_scenes}. Details of each task can be found in Appendix A.  
% \VK{What is a task? Do we have a list of all 10 tasks. What do you mean by environment here?}. [TODO: Zoey fill in Vikash's questions]

\begin{table*}[!h]
\centering
\caption{Real-World Robot Experiments tested on 10 tasks. On average, \name{} achieves $85\%$ success rate on unseen environment, $52\%$ on unseen object to place, and $45\%$ on unseen object to pick.}
\vspace{-0.3em}
\label{table:robot_experiment}
\begin{tabular}{P{2.4cm}|P{0.8cm}|P{0.7cm}|P{1.2cm}|P{1.2cm}|P{0.8cm}|P{0.7cm}|P{1.4cm}|P{1.4cm}|P{1.2cm}|P{1.4cm}} 
\hline
                   & \begin{tabular}[c]{@{}c@{}}bowl to \\ Coaster\end{tabular} & \begin{tabular}[c]{@{}c@{}}box to \\ basket\end{tabular} & \begin{tabular}[c]{@{}c@{}}red bowl to \\ blue bowl\end{tabular} & \begin{tabular}[c]{@{}c@{}}red plate to \\ tray\end{tabular} & \begin{tabular}[c]{@{}c@{}}box to \\ tray\end{tabular} & \begin{tabular}[c]{@{}c@{}}plate to \\ box\end{tabular} & \begin{tabular}[c]{@{}c@{}}white plate to \\ blue plate\end{tabular} & \begin{tabular}[c]{@{}c@{}}coaster to \\ salt container\end{tabular} & \begin{tabular}[c]{@{}c@{}}coaster to \\ dust bin\end{tabular} & \begin{tabular}[c]{@{}c@{}}mouse box to \\ fruit box\end{tabular} \\ \hline
Unseen Environment & 0.8                                                        & 0.9                                                      & 1                                                                & 1                                                            & 1                                                      & 0.9                                                     & 0.9                                                                  & 1                                                                    & 0.5                                                            & 0.5                                                               \\ \hline
Unseen Place       & 0.7                                                        & 1                                                        & 0.5                                                              & 0.3                                                          & 0.6                                                    & 0.3                                                     & 0.4                                                                  & 0.4                                                                  & 0.4                                                            & 0.6                                                               \\ \hline
Unseen Pick        & 0.2                                                        & 0.6                                                      & 0.5                                                              & 0.6                                                          & 0.7                                                    & 0.3                                                     & 0.3                                                                  & 0.7                                                                  & 0                                                              & 0.6                                                               \\ \hline

\end{tabular}
\end{table*}

\begin{table*}
\centering
\caption{Evaluating with and without \name{} on unseen scenes collected in the real world across 10 tasks. On average, \name{} shows significant improvement in unseen environments and objects.}
\vspace{-0.3em}
\label{table:robot_baselines}
\begin{tabular}{|P{2cm}|P{0.1cm}|P{0.1cm}|P{0.5cm}|P{0.1cm}|P{0.1cm}|P{0.5cm}|P{0.1cm}|P{0.1cm}|P{0.5cm}|P{0.1cm}|P{0.1cm}|P{0.5cm}|P{0.1cm}|P{0.1cm}|P{0.5cm}|}
\hline
\textbf{}     & \multicolumn{3}{c|}{\textbf{box to tray}}                    & \multicolumn{3}{c|}{\textbf{box to basket}}                  & \multicolumn{3}{c|}{\textbf{coaster to  dust pan}}           & \multicolumn{3}{c|}{\textbf{plate  to tray}}                 & \multicolumn{3}{c|}{\textbf{bowl to coaster}}                \\ \hline
              & \multicolumn{1}{c|}{env} & \multicolumn{1}{c|}{pick} & place & \multicolumn{1}{c|}{env} & \multicolumn{1}{c|}{pick} & place & \multicolumn{1}{c|}{env} & \multicolumn{1}{c|}{pick} & place & \multicolumn{1}{c|}{env} & \multicolumn{1}{c|}{pick} & place & \multicolumn{1}{c|}{env} & \multicolumn{1}{c|}{pick} & place \\ \hline
No GenAug     & \multicolumn{1}{c|}{0.8} & \multicolumn{1}{c|}{0}    & 0     & \multicolumn{1}{c|}{0.2} & \multicolumn{1}{c|}{0.2}  & 0     & \multicolumn{1}{c|}{0.8} & \multicolumn{1}{c|}{0.4}  & 0.4   & \multicolumn{1}{c|}{0}   & \multicolumn{1}{c|}{0}    & 0     & \multicolumn{1}{c|}{0}   & \multicolumn{1}{c|}{0}    & 0     \\ \hline
GenAug        & \multicolumn{1}{c|}{1}   & \multicolumn{1}{c|}{0.6}  & 1     & \multicolumn{1}{c|}{0.6} & \multicolumn{1}{c|}{0.6}  & 0.8   & \multicolumn{1}{c|}{1}   & \multicolumn{1}{c|}{0.4}  & 0.4   & \multicolumn{1}{c|}{1}   & \multicolumn{1}{c|}{0.4}  & 0.2   & \multicolumn{1}{c|}{0.6} & \multicolumn{1}{c|}{0.6}  & 0.6   \\ \hline
\textbf{}     & \multicolumn{3}{c|}{\textbf{plate to plate}}                 & \multicolumn{3}{c|}{\textbf{box to box}}                     & \multicolumn{3}{c|}{\textbf{plate to box}}                   & \multicolumn{3}{c|}{\textbf{coaster to salt}}                & \multicolumn{3}{c|}{\textbf{bowl to bowl}}                   \\ \hline
              & \multicolumn{1}{c|}{env} & \multicolumn{1}{c|}{pick} & place & \multicolumn{1}{c|}{env} & \multicolumn{1}{c|}{pick} & place & \multicolumn{1}{c|}{env} & \multicolumn{1}{c|}{pick} & place & \multicolumn{1}{c|}{env} & \multicolumn{1}{c|}{pick} & place & \multicolumn{1}{c|}{env} & \multicolumn{1}{c|}{pick} & place \\ \hline
No GenAug  & \multicolumn{1}{c|}{0}   & \multicolumn{1}{c|}{0}    & 0.2   & \multicolumn{1}{c|}{0.2} & \multicolumn{1}{c|}{0}    & 0     & \multicolumn{1}{c|}{0.6} & \multicolumn{1}{c|}{0.2}  & 0     & \multicolumn{1}{c|}{0.2} & \multicolumn{1}{c|}{0}    & 0.2   & \multicolumn{1}{c|}{1}   & \multicolumn{1}{c|}{0.2}  & 0     \\ \hline
GenAug     & \multicolumn{1}{c|}{1}   & \multicolumn{1}{c|}{0}    & 0.6   & \multicolumn{1}{c|}{0.8} & \multicolumn{1}{c|}{0.4}  & 0.4   & \multicolumn{1}{c|}{1}   & \multicolumn{1}{c|}{0.8}  & 0     & \multicolumn{1}{c|}{1}   & \multicolumn{1}{c|}{0.4}  & 0.4   & \multicolumn{1}{c|}{1}   & \multicolumn{1}{c|}{0.4}  & 1     \\ \hline
\end{tabular}
\vspace{-1em}
\end{table*}

\subsection{Augmentation Infrastructure}
\label{sec:system_genaug}
As described in Section \ref{sec: system_setup}, \name{} requires object meshes to generate cross-category augmentations and distractors. To perform this augmentation, we use 40 object meshes from the GoogleScan dataset \cite{downs2022google} and Free3D \cite{free3d}. Of these, we choose 11 objects to augment the original object of interest and 12 objects to augment the target receptacle. Any of the remaining 38 objects are then randomly chosen as distractors. During augmentation, we randomly select which components (table, object texture, shape, distractors) to change to generate the augmented training dataset. For each demonstration, we apply \name{} 100 times resulting in 1000 augmented environments per task. The augmented data is then passed into Cliport \cite{shridhar2021cliport} to learn a language-conditioned policy.

\section{Experiment}
\label{sec:experiment}
We evaluate the effectiveness of \name{} in both the real world and simulation. Our goal is to: (1) demonstrate \name{} is practical and effective for real-world robot learning, (2) compare \name{} with other baselines in end-to-end vision manipulation tasks, (3) investigate different design choices in \name{}. We will first show our results in a real-world setting, followed by an in-depth baseline study in simulation.

\subsection{Real-world Experiment}
% \VK{we also have r3m.CLIPort We are evaluating in situations more than just augmentations} 
% (1) Can funAug provide diverse and realistic augmentation on the real-world dataset? \VK{not quantifiable. I'd recommend starting with the central pitch of the paper i.e. the main result}

% (2) Does the resulting augmented dataset make robot more robust at unseen scenes? How does this compare with existing augmentation methods?

% (3) Can we train a robot policy with FunAug and generalize to unseen environments? 
\subsubsection{Design of demo and test environment}
\begin{figure}[!h]
\vspace{-0.5em}
    \centering
    \includegraphics[width=0.49\textwidth]{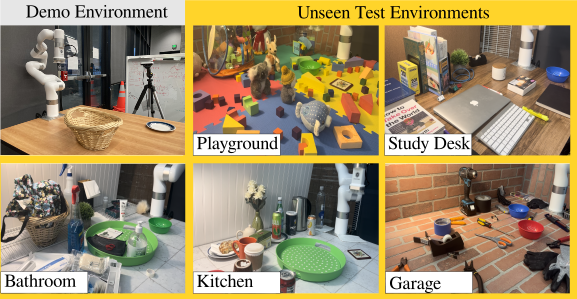}
    \caption{Demonstration environment and examples of test environments used in our robot experiments.}
    \vspace{-0.5em}
    \label{fig:test_scenes}
\end{figure}

To show the generalization capability of a model trained with \name{}, we collect demonstrations of 10 tasks in one single environment and create different styles of test environments such as "Playground", "Study Desk", "Kitchen Island", "Garage" and "Bathroom" as shown in Figure \ref{fig:test_scenes}. For evaluation, we randomly add and rearrange objects from each test style and create unseen environments. Please see Appendix A for further details. 
\subsubsection{Result}
We train CLIPort with augmented RGBD and text prompts for tasks collected in the real world and evaluate in various unseen environments. In particular, for each task, we randomly choose an environment style from Figure \ref{fig:test_scenes}, randomly rearrange and add objects on the table to create 10 unseen environments, 10 scenes with unseen objects to pick, and 10 scenes with unseen objects to place. We observe that \name{} shows a significant generalization to unseen environments with an average of $85\%$ success rate. On more challenging tasks of unseen objects to pick or place, \name{} is able to achieve $45\%$ and $52\%$ success rates, which are expected to improve with more demonstrations and more object meshes for augmentation.
% The performance drops for unseen objects to place or pick, but \name{} is still able to achieve the success of $52\%$ and $45\%$. 
Results for each task can be found in Table \ref{table:robot_experiment}. 
\begin{figure*}[!h]
    \centering
    \includegraphics[width=\textwidth]{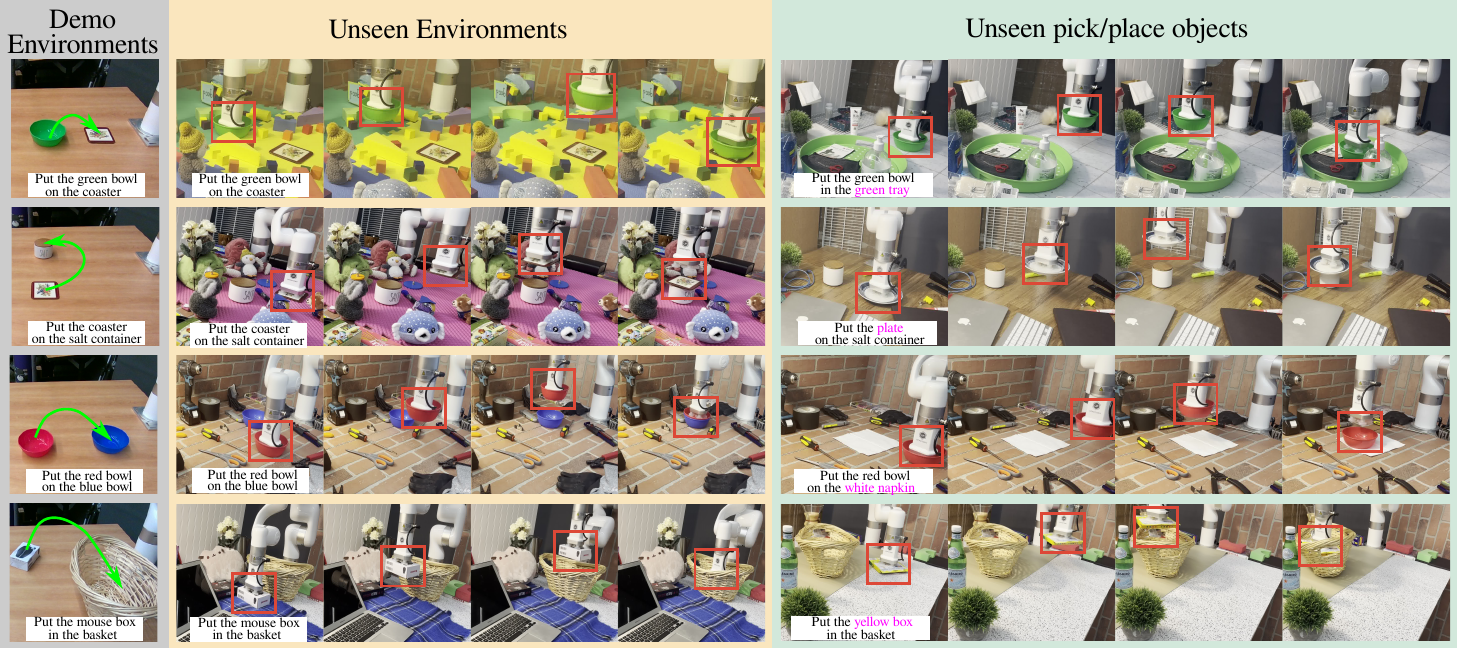}
    \vspace{-1.5em}\caption{Examples of real-world experiments. Given demonstrations in one simple environment, GenAug enables the robot to generalize unseen environments and objects. }
    \vspace{-1.5em}
    \label{fig:robot_task}
\end{figure*}
\subsubsection{Baselines for real-world experiments}
To further show the effectiveness of \name{}, we compare our approach with CLIPort trained without \name{}, shown in Table \ref{table:robot_baselines}. To ensure both methods are tested with the same input observations, we evaluate the success rate by comparing the predicted pick and place affordances with ground truth locations. For each task, we evaluate both methods on 5 unseen environments, 5 unseen objects to pick, and 5 unseen objects to place.
By averaging the success rates from Table \ref{table:robot_baselines}, we observe \name{} provides a significant improvement for zero-shot generalization. In particular, \name{} achieves $80\%$ success rate on unseen environments compared to $38\%$ without \name{}. On unseen objects to place, \name{} achieves $54\%$ success rate compared to $8\%$ without. Finally, \name{} achieves $46\%$ success rate on unseen objects to pick compared to $10\%$ without. We visualize and compare the differences in their predicted affordances in Appendix A.

% We observe \name{} provides a significant improvement for zero-shot generalization compared to training without \name{}, with $42\%$ improvement on "unseen environment", $36\%$ improvement on "unseen object to pick", and $46\%$ improvement on "unseen object to place". We visualize and compare the differences in their predicted affordances in Appendix A.

\subsection{Simulation}
To further study in depth the effectiveness of \name{}, we conduct large-scale experiments with other baselines in simulation. In particular, we organize baseline methods as (1) in-domain augmentation methods and (2) learning from out-of-domain priors, as described below.

\subsubsection{In-domain augmentation methods}
(1) "No Augmentation" does not use any data augmentation techniques. 
(2) "Spatial Augmentation" randomly transforms the cropped object image features to learn rotation and translation equivariance, as introduced in TransporterNet \cite{zeng2020transporter}. 
(3)"Random Copy Paste" randomly queries objects and their segmented images from LVIS dataset \cite{gupta2019lvis}, and places them in the original scene. This includes adding distractors around the pick or place objects or replacing them. Further visualization of this approach can be found in Appendix C. 
(4)"Random Background" does not modify the pick or place objects but replaces the table and background with images randomly selected from LVIS dataset.
(5)"Random Distractors" randomly selects segmented images from LVIS dataset as distractors.

\subsubsection{ Learning from out-of-domain priors}
In addition, we investigate whether learning from a pretrained out-of-domain visual representation can improve the zero-shot capability on challenging unseen environments. In particular, we initialize the network with pre-trained R3M \cite{nair2022r3m} weights and finetune it on our dataset. 

We use baselines described above with two imitation learning methods: TransportNet \cite{zeng2020transporter} and CLIPort \cite{shridhar2021cliport}. Since all the baselines cannot update the depth of the augmented images, we only use RGB images instead of RGBD used in the original TransporterNet and CLIPort. For each baseline, we train 5 tasks in simulation and report their average success rate in Table \ref{table:sim_experiment}. We observe \name{} significantly outperforms other approaches in most of the tasks. One interesting observation is that randomly copying and pasting segmented images or replacing the background images can provide reasonable robustness in unseen environments but are not able to achieve similar performance as \name{} at unseen objects. This indicates generating physically plausible scenes that are semantically meaningful is important. Details of tasks in simulation experiments can be found in Appendix B. 

\begin{table*}
\centering
\caption{Baseline experiments evaluated in simulation. We compare the average  performance of \name{} with other methods on 5 pick-and-place tasks and observe \name{} provides a significant improvement at unseen environments and objects.}
\vspace{-0.5em}
\label{table:sim_experiment}
\begin{tabular}{P{2.5cm}P{0.4cm}P{0.4cm}P{0.4cm}P{0.4cm}P{0.4cm}P{0.4cm}P{0.4cm}P{0.4cm}P{0.4cm}P{0.4cm}P{0.4cm}P{0.4cm}P{0.4cm}P{0.4cm}P{0.4cm}P{0.4cm}P{0.4cm}P{0.4cm}} 
\multicolumn{1}{l}{}      & \multicolumn{6}{c}{Unseen Environment}                                                        & \multicolumn{6}{c}{Unseen to place}                                                           & \multicolumn{6}{c}{Unseen to pick}                                                            \\ \cline{2-19} 
\multicolumn{1}{l}{}      & \multicolumn{3}{c}{TransporterNet}            & \multicolumn{3}{c}{CLIPort}                   & \multicolumn{3}{c}{TransporterNet}            & \multicolumn{3}{c}{CLIPort}                   & \multicolumn{3}{c}{TransporterNet}            & \multicolumn{3}{c}{CLIPort}                   \\ \cline{2-19} 
Method                    & 1             & 10            & 100           & 1             & 10            & 100           & 1             & 10            & 100           & 1             & 10            & 100           & 1             & 10            & 100           & 1             & 10            & 100           \\ \cline{2-19} 
No Augmentation           & 4.8           & 8.1           & 9.8           & 11.7          & 14.3          & 14.4          & 15.1          & 30.4          & 52.6          & 39.4          & 40.8          & 44.6          & 8.5           & 34.6          & 54.9          & 46.0          & 67.0          & 64.1          \\
Spatial Augmentation      & 11.0          & 12.2          & 8.3           & 23.3          & 16.1          & 26.7          & 44.3          & 50.5          & 65.3          & 26.1          & 36.9          & 50.7          & 53.6          & 57.2          & 66.4          & 38.2          & 56.9          & 80.3          \\
Random Copy Paste         & \textbf{53.1} & 67.0          & 73.5          & 38.2          & 39.8          & 64.3          & 55.1          & 65.4          & \textbf{84.9} & 39.7          & 55.9          & 73.9          & 48.3          & 67.0          & 76.1          & 52.5          & 65.0          & 81.0          \\
Random Background         & 53.0          & 75.3          & 79.1          & 33.6          & 62.2          & 55.4          & 24.5          & 22.1          & 35.5          & 7.6           & 9.9           & 17.9          & 44.4          & 40.7          & 35.9          & 19.2          & 52.7          & 72.3          \\
Random Distractors        & 10.1          & 9.7           & 13.7          & 15.4          & 36.2          & 35.8          & 28.2          & 60.7          & 66.0          & 27.5          & 51.8          & 54.3          & 42.5          & 47.4          & 62.3          & 31.0          & 64.0          & 69.1          \\
R3M Finetune              & 4.1           & 6.0           & 4.8           & 22.2          & 16.8          & 20.9          & 43.5          & 40.6          & 41.9          & 30.9          & 43.5          & 57.5          & 45.6          & 45.7          & 41.1          & 46.7          & 50.7          & 72.7          \\
\textbf{\name{}}           & 43.9          & 58.5          & 77.6          & 46.6          & 57.0          & 71.9          & \textbf{69.1} & \textbf{76.5} &  83.6    & 62.6          & \textbf{83.9} & \textbf{86.3} & \textbf{75.3} & 75.6          & \textbf{87.2} & \textbf{61.5} & \textbf{77.7} & \textbf{83.1} \\
\textbf{\name{} (w Depth)} & 47.8          & \textbf{83.8} & \textbf{91.2} & \textbf{47.2} & \textbf{60.9} & \textbf{73.4} & 39.9          & 67.2          & 74.2          & \textbf{64.8} & 73.8          & 84.6          & 71.2          & \textbf{83.4} & 87.1          & 56.2          & 67.3          & 81.5         
\end{tabular}
\vspace{-2em}
\end{table*}

\subsection{Ablations}
In this section, we aim to study different design choices in \name{}. In particular, our goal is to investigate (1) how the number of augmentations affects the generalization performance to unseen environments, (2) when \name{} will fail in real-world unseen environments. We further justify the choice of using depth-guided models and compare this with in-painting models in Appendix A. 
\begin{figure}[!h]
\vspace{-0.5em}
\centering
\includegraphics[width=0.49\textwidth]{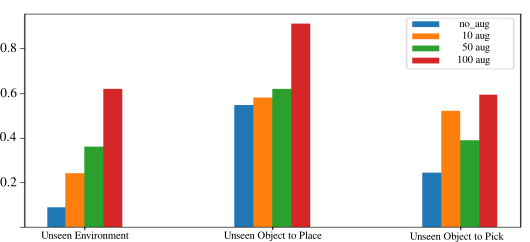}
\caption{Analysis of the number of augmentation on unseen scenes.}
\vspace{-0.5em}
\label{fig:ablation_aug}
\end{figure}

\subsubsection{Impact of the number of augmentations} Given the task "put the brown plate in the brown box" in simulation, we apply \name{} $0, 10, 50 $ and $100$ times and compare their success rate on 100 scenes of "unseen environment", 100 scenes of "unseen object to pick" and 100 scenes of "unseen object to place". As shown in Figure \ref{fig:ablation_aug}, the performance improves as the number of augmentations increases, which indicates the importance of using augmentation as a way to robustify the generalization capability. 
% \begin{figure}[!h]
% \vspace{-0.1em}
% \centering
% \includegraphics[width=0.49\textwidth]{figures/failure_mode.png}
% \caption{Failure cases observed in real-world experiments. Ground truth actions are in green while predicted ones are in red.}
% \vspace{-1em}
% \label{fig:failures}
% \end{figure}

\subsubsection{Failure cases} In addition, we also analyze failure cases and visualize them in Appendix A. We observe failure cases usually happen when the object and background share similar colors or the table is cluttered with too many objects.

\section{Related Work}
% We organize our related works by techniques bringing structural invariances and priors to aid robot learning.

\paragraph{\textbf{Image space Augmentation}} In the absence of diverse data, a promising direction is finding ways to inject structure directly into learned models for widespread generalization. The most widely used technique is various forms of data augmentation ~\cite{shorten19aug}, such as cropping, shifting, noise injection and rotation. These methods have been used in many robot learning approaches and provide a significant improvement in data efficiency \cite{benton2020learning, cubuk2018autoaugment, shorten2019survey, kostrikov2020image}. For example, \cite{zafar2022comparison} investigate different augmentation modes in Meta-learning settings. In addition, several methods attempt to enforce geometric invariance through architectural innovations such as ~\cite{wang22equivariant} and \cite{deng2021vector}. 
While these methods can provide a local notion of robustness and invariance to perceptual noise, they do not provide generalization to novel object shapes or scenes. More recently out-of-domain models have started making their way into robot learning. For example - \cite{kapelyukh2022dall} uses large text-image models like Dall-e \cite{ramesh2022hierarchical} to generate favorable image goals for robots, and CACTI \cite{mandi2022cacti} shows adding distractor objects using in-painting models \cite{rombach2022high} improves multi-task policies. 
These approaches while helpful in task specification, provide limited benefit for robots to generalize to entirely unseen situations. In contrast, \name{} induces semantic changes to the observations thereby helping acquire behavior invariance to new scenes.

\paragraph{\textbf{Leveraging Simulation Data}}
A natural solution to this problem is to leverage simulation data ~\cite{deitke22procthor}. However, while this method is efficient at generating a large \emph{quantity} of data, it can be challenging to create diverse content (objects meshes, physics, layouts, and visual appearances), and often the resulting simulations exhibit a significant gap from reality. To address this issue, domain randomization \cite{tobin2017domain, tremblay2018training} varies multiple simulation parameters in order to boost the effectiveness of trained policies in the real world. Randomization of lighting and camera parameters during training can allow a policy to be invariant to the effects of lighting and visual perturbations in the real world. Physics parameters of the scene can also be randomized to transfer policies trained in simulation to the real world \cite{tobin2018domain}. While effective, challenges in creating a simulation with visual and physical realism for every task and behavior severely restrict the applicability of these methods to isolated known tasks and limited diversity. Furthermore, they require the user to define augmentation parameters and their ranges which can be very nontrivial for complex tasks \cite{akkaya2019solving, handa2022dextreme}.

% In contrast to all the methods above, \name{} can be directly applied to real robot experiences. Text-to-image models trained on internet scale data bring in geometric and contextural invariances at no additional cost.

% \name{}, in contrast, aids agents to generalize their behavior to new settings by generating augmented experiences for robots to learn from. 
% Another group of methods leverages large visual models trained on out-of-domain datasets to bring visual priors to aid with robot learning \cite{nair2022r3m, radosavovic2022real, parisi2022unsurprising}. The visual inductive biases acquired by these models from web-scale data help policies focus on task-relevant semantically meaningful features. Resulting policies develop visual invariance towards details not important for the task such as lighting and texture. These methods are complementary to \name{} which delivers semantic augmentations of robot experiences in a new context. These augmented experiences can be used in conjunction with the out-of-domain visual methods to deliver semantic as well as a visual generalizations in robot behaviors. 
\section{Limitations}
\label{sec:limitations}
% \begin{enumerate} 
\textbf{Action Assumption: } Despite showing promising visual diversity, \name{} does not augment action labels and reason about physics parameters such as material, friction, or deformation, thus it assumes the same action still works on the augmented scenes. For the augmented cluttered scenes, \name{} assumes the same action trajectory is not colliding with the augmented objects. 

\textbf{Augmentation and Speed: } \name{} cannot guarantee visual consistency for frame augmentation in a video. \name{} usually takes about 30 seconds to complete all the augmentations for one scene, which might not be practical for some robot learning approaches such as on-policy RL. 
% Visual effect: shadows, lightings, bad generations, etc
% \end{enumerate}
\section{Conclusion}
We present \name{}, a novel system for augmenting real-world robot data. \name{} leverages a data augmentation approach that bootstraps a small number of human demonstrations into a large dataset with diverse and novel objects. We demonstrate that \name{} is able to train a robot that generalizes to entirely unseen environments and objects, with $40\%$ improvement over training without \name{}. For future work, we hope to investigate \name{} in other domains of robot learning such as Behavior Cloning and Reinforcement learning, and extend our table-top tasks into more challenging manipulation problems. Moreover, investigating whether a combination of language models and vision-language models can yield impressive scene generations would be a promising direction in the future. 
\section{Acknowledgement}
We thank Aaron Walsman for helping with transporting materials from Home Depot for creating robot environments. We thank Mohit Shridhar for the discussions about CLIPort training. We thank all members from the WEIRD lab, the RSE lab at the University of Washington, as well as Kay Ke, Yunchu Zhang, and Abhay Deshpande for many discussions, support, late-night food ordering, and snack sharing. We are thankful to Boling Yang and Henri Fung for their patience in letting us book all the slots for using the machine before the deadline. We are also grateful to Selest Nashef for his help to keep our robot experiments safe and organized. Part of this work was done while Zoey Chen was an Intern at Meta AI. The work was also
funded in part by Amazon Science Hub.

\bibliographystyle{unsrtnat}
\bibliography{main.bib}

\begin{thebibliography}{38}
\providecommand{\natexlab}[1]{#1}
\providecommand{\url}[1]{\texttt{#1}}
\expandafter\ifx\csname urlstyle\endcsname\relax
  \providecommand{\doi}[1]{doi: #1}\else
  \providecommand{\doi}{doi: \begingroup \urlstyle{rm}\Url}\fi

\bibitem[Levine et~al.(2015)Levine, Finn, Darrell, and
  Abbeel]{levine15endtoend}
Sergey Levine, Chelsea Finn, Trevor Darrell, and Pieter Abbeel.
\newblock End-to-end training of deep visuomotor policies.
\newblock \emph{CoRR}, abs/1504.00702, 2015.
\newblock URL \url{http://arxiv.org/abs/1504.00702}.

\bibitem[Nagabandi et~al.(2019)Nagabandi, Konoglie, Levine, and
  Kumar]{nagabandi:corl2019}
Anusha Nagabandi, Kurt Konoglie, Sergey Levine, and Vikash Kumar.
\newblock Deep dynamics models for learning dexterous manipulation.
\newblock In \emph{CoRL}, 2019.

\bibitem[Qureshi et~al.(2018)Qureshi, Bency, and Yip]{qureshi18mpn}
Ahmed~Hussain Qureshi, Mayur~J. Bency, and Michael~C. Yip.
\newblock Motion planning networks.
\newblock \emph{CoRR}, abs/1806.05767, 2018.
\newblock URL \url{http://arxiv.org/abs/1806.05767}.

\bibitem[Fishman et~al.(2022)Fishman, Murali, Eppner, Peele, Boots, and
  Fox]{fishman22mpn}
Adam Fishman, Adithyavairavan Murali, Clemens Eppner, Bryan Peele, Byron Boots,
  and Dieter Fox.
\newblock Motion policy networks.
\newblock \emph{CoRR}, abs/2210.12209, 2022.
\newblock \doi{10.48550/arXiv.2210.12209}.
\newblock URL \url{https://doi.org/10.48550/arXiv.2210.12209}.

\bibitem[Kalashnikov et~al.(2018)Kalashnikov, Irpan, Pastor, Ibarz, Herzog,
  Jang, Quillen, Holly, Kalakrishnan, Vanhoucke, and
  Levine]{kalashnikov18qtopt}
Dmitry Kalashnikov, Alex Irpan, Peter Pastor, Julian Ibarz, Alexander Herzog,
  Eric Jang, Deirdre Quillen, Ethan Holly, Mrinal Kalakrishnan, Vincent
  Vanhoucke, and Sergey Levine.
\newblock Qt-opt: Scalable deep reinforcement learning for vision-based robotic
  manipulation.
\newblock \emph{CoRR}, abs/1806.10293, 2018.
\newblock URL \url{http://arxiv.org/abs/1806.10293}.

\bibitem[Brohan et~al.(2022)Brohan, Brown, Carbajal, Chebotar, Dabis, Finn,
  Gopalakrishnan, Hausman, Herzog, Hsu, Ibarz, Ichter, Irpan, Jackson,
  Jesmonth, Joshi, Julian, Kalashnikov, Kuang, Leal, Lee, Levine, Lu, Malla,
  Manjunath, Mordatch, Nachum, Parada, Peralta, Perez, Pertsch, Quiambao, Rao,
  Ryoo, Salazar, Sanketi, Sayed, Singh, Sontakke, Stone, Tan, Tran, Vanhoucke,
  Vega, Vuong, Xia, Xiao, Xu, Xu, Yu, and Zitkovich]{brohan23rt1}
Anthony Brohan, Noah Brown, Justice Carbajal, Yevgen Chebotar, Joseph Dabis,
  Chelsea Finn, Keerthana Gopalakrishnan, Karol Hausman, Alexander Herzog,
  Jasmine Hsu, Julian Ibarz, Brian Ichter, Alex Irpan, Tomas Jackson, Sally
  Jesmonth, Nikhil~J. Joshi, Ryan Julian, Dmitry Kalashnikov, Yuheng Kuang,
  Isabel Leal, Kuang{-}Huei Lee, Sergey Levine, Yao Lu, Utsav Malla, Deeksha
  Manjunath, Igor Mordatch, Ofir Nachum, Carolina Parada, Jodilyn Peralta,
  Emily Perez, Karl Pertsch, Jornell Quiambao, Kanishka Rao, Michael~S. Ryoo,
  Grecia Salazar, Pannag Sanketi, Kevin Sayed, Jaspiar Singh, Sumedh Sontakke,
  Austin Stone, Clayton Tan, Huong Tran, Vincent Vanhoucke, Steve Vega, Quan
  Vuong, Fei Xia, Ted Xiao, Peng Xu, Sichun Xu, Tianhe Yu, and Brianna
  Zitkovich.
\newblock {RT-1:} robotics transformer for real-world control at scale.
\newblock \emph{CoRR}, abs/2212.06817, 2022.
\newblock \doi{10.48550/arXiv.2212.06817}.
\newblock URL \url{https://doi.org/10.48550/arXiv.2212.06817}.

\bibitem[Mahler et~al.(2017)Mahler, Liang, Niyaz, Laskey, Doan, Liu, Ojea, and
  Goldberg]{mahler17dexnet}
Jeffrey Mahler, Jacky Liang, Sherdil Niyaz, Michael Laskey, Richard Doan, Xinyu
  Liu, Juan~Aparicio Ojea, and Ken Goldberg.
\newblock Dex-net 2.0: Deep learning to plan robust grasps with synthetic point
  clouds and analytic grasp metrics.
\newblock In Nancy~M. Amato, Siddhartha~S. Srinivasa, Nora Ayanian, and Scott
  Kuindersma, editors, \emph{Robotics: Science and Systems XIII, Massachusetts
  Institute of Technology, Cambridge, Massachusetts, USA, July 12-16, 2017},
  2017.
\newblock \doi{10.15607/RSS.2017.XIII.058}.
\newblock URL \url{http://www.roboticsproceedings.org/rss13/p58.html}.

\bibitem[Pomerleau(1988)]{pomerleau88alvinn}
Dean Pomerleau.
\newblock {ALVINN:} an autonomous land vehicle in a neural network.
\newblock In David~S. Touretzky, editor, \emph{Advances in Neural Information
  Processing Systems 1, {[NIPS} Conference, Denver, Colorado, USA, 1988]},
  pages 305--313. Morgan Kaufmann, 1988.
\newblock URL
  \url{http://papers.nips.cc/paper/95-alvinn-an-autonomous-land-vehicle-in-a-neural-network}.

\bibitem[Schuhmann et~al.(2022)Schuhmann, Beaumont, Vencu, Gordon, Wightman,
  Cherti, Coombes, Katta, Mullis, Wortsman, Schramowski, Kundurthy, Crowson,
  Schmidt, Kaczmarczyk, and Jitsev]{schuhmann22laion}
Christoph Schuhmann, Romain Beaumont, Richard Vencu, Cade Gordon, Ross
  Wightman, Mehdi Cherti, Theo Coombes, Aarush Katta, Clayton Mullis, Mitchell
  Wortsman, Patrick Schramowski, Srivatsa Kundurthy, Katherine Crowson, Ludwig
  Schmidt, Robert Kaczmarczyk, and Jenia Jitsev.
\newblock {LAION-5B:} an open large-scale dataset for training next generation
  image-text models.
\newblock \emph{CoRR}, abs/2210.08402, 2022.
\newblock \doi{10.48550/arXiv.2210.08402}.
\newblock URL \url{https://doi.org/10.48550/arXiv.2210.08402}.

\bibitem[Deng et~al.(2009)Deng, Dong, Socher, Li, Li, and
  Fei{-}Fei]{deng09imagenet}
Jia Deng, Wei Dong, Richard Socher, Li{-}Jia Li, Kai Li, and Li~Fei{-}Fei.
\newblock Imagenet: {A} large-scale hierarchical image database.
\newblock In \emph{2009 {IEEE} Computer Society Conference on Computer Vision
  and Pattern Recognition {(CVPR} 2009), 20-25 June 2009, Miami, Florida,
  {USA}}, pages 248--255. {IEEE} Computer Society, 2009.
\newblock \doi{10.1109/CVPR.2009.5206848}.
\newblock URL \url{https://doi.org/10.1109/CVPR.2009.5206848}.

\bibitem[Shridhar et~al.(2021)Shridhar, Manuelli, and Fox]{shridhar2021cliport}
Mohit Shridhar, Lucas Manuelli, and Dieter Fox.
\newblock Cliport: What and where pathways for robotic manipulation.
\newblock In \emph{Proceedings of the 5th Conference on Robot Learning (CoRL)},
  2021.

\bibitem[Zeng et~al.(2020)Zeng, Florence, Tompson, Welker, Chien, Attarian,
  Armstrong, Krasin, Duong, Sindhwani, et~al.]{zeng2020transporter}
Andy Zeng, Pete Florence, Jonathan Tompson, Stefan Welker, Jonathan Chien,
  Maria Attarian, Travis Armstrong, Ivan Krasin, Dan Duong, Vikas Sindhwani,
  et~al.
\newblock Transporter networks: Rearranging the visual world for robotic
  manipulation.
\newblock \emph{arXiv preprint arXiv:2010.14406}, 2020.

\bibitem[Benton et~al.(2020)Benton, Finzi, Izmailov, and
  Wilson]{benton2020learning}
Gregory Benton, Marc Finzi, Pavel Izmailov, and Andrew~G Wilson.
\newblock Learning invariances in neural networks from training data.
\newblock \emph{Advances in Neural Information Processing Systems},
  33:\penalty0 17605--17616, 2020.

\bibitem[Cubuk et~al.(2018)Cubuk, Zoph, Mane, Vasudevan, and
  Le]{cubuk2018autoaugment}
Ekin~D Cubuk, Barret Zoph, Dandelion Mane, Vijay Vasudevan, and Quoc~V Le.
\newblock Autoaugment: Learning augmentation policies from data.
\newblock \emph{arXiv preprint arXiv:1805.09501}, 2018.

\bibitem[Shorten and Khoshgoftaar(2019{\natexlab{a}})]{shorten2019survey}
Connor Shorten and Taghi~M Khoshgoftaar.
\newblock A survey on image data augmentation for deep learning.
\newblock \emph{Journal of big data}, 6\penalty0 (1):\penalty0 1--48,
  2019{\natexlab{a}}.

\bibitem[Perez and Wang(2017)]{perez2017effectiveness}
Luis Perez and Jason Wang.
\newblock The effectiveness of data augmentation in image classification using
  deep learning.
\newblock \emph{arXiv preprint arXiv:1712.04621}, 2017.

\bibitem[Singer et~al.(2022)Singer, Polyak, Hayes, Yin, An, Zhang, Hu, Yang,
  Ashual, Gafni, et~al.]{singer2022make}
Uriel Singer, Adam Polyak, Thomas Hayes, Xi~Yin, Jie An, Songyang Zhang, Qiyuan
  Hu, Harry Yang, Oron Ashual, Oran Gafni, et~al.
\newblock Make-a-video: Text-to-video generation without text-video data.
\newblock \emph{arXiv preprint arXiv:2209.14792}, 2022.

\bibitem[Ramesh et~al.(2022)Ramesh, Dhariwal, Nichol, Chu, and
  Chen]{ramesh2022hierarchical}
Aditya Ramesh, Prafulla Dhariwal, Alex Nichol, Casey Chu, and Mark Chen.
\newblock Hierarchical text-conditional image generation with clip latents.
\newblock \emph{arXiv preprint arXiv:2204.06125}, 2022.

\bibitem[Rombach et~al.(2022)Rombach, Blattmann, Lorenz, Esser, and
  Ommer]{rombach2022high}
Robin Rombach, Andreas Blattmann, Dominik Lorenz, Patrick Esser, and Bj{\"o}rn
  Ommer.
\newblock High-resolution image synthesis with latent diffusion models.
\newblock In \emph{Proceedings of the IEEE/CVF Conference on Computer Vision
  and Pattern Recognition}, pages 10684--10695, 2022.

\bibitem[Saharia et~al.(2022)Saharia, Chan, Saxena, Li, Whang, Denton,
  Ghasemipour, Ayan, Mahdavi, Lopes, et~al.]{saharia2022photorealistic}
Chitwan Saharia, William Chan, Saurabh Saxena, Lala Li, Jay Whang, Emily
  Denton, Seyed Kamyar~Seyed Ghasemipour, Burcu~Karagol Ayan, S~Sara Mahdavi,
  Rapha~Gontijo Lopes, et~al.
\newblock Photorealistic text-to-image diffusion models with deep language
  understanding.
\newblock \emph{arXiv preprint arXiv:2205.11487}, 2022.

\bibitem[Rombach et~al.(2021)Rombach, Blattmann, Lorenz, Esser, and
  Ommer]{rombach2021highresolution}
Robin Rombach, Andreas Blattmann, Dominik Lorenz, Patrick Esser, and Björn
  Ommer.
\newblock High-resolution image synthesis with latent diffusion models, 2021.

\bibitem[Downs et~al.(2022)Downs, Francis, Koenig, Kinman, Hickman, Reymann,
  McHugh, and Vanhoucke]{downs2022google}
Laura Downs, Anthony Francis, Nate Koenig, Brandon Kinman, Ryan Hickman, Krista
  Reymann, Thomas~B McHugh, and Vincent Vanhoucke.
\newblock Google scanned objects: A high-quality dataset of 3d scanned
  household items.
\newblock \emph{arXiv preprint arXiv:2204.11918}, 2022.

\bibitem[free3d()]{free3d}
free3d.
\newblock {free3d}.
\newblock \url{https://free3d.com/}.

\bibitem[Gupta et~al.(2019)Gupta, Dollar, and Girshick]{gupta2019lvis}
Agrim Gupta, Piotr Dollar, and Ross Girshick.
\newblock {LVIS}: A dataset for large vocabulary instance segmentation.
\newblock In \emph{Proceedings of the {IEEE} Conference on Computer Vision and
  Pattern Recognition}, 2019.

\bibitem[Nair et~al.(2022)Nair, Rajeswaran, Kumar, Finn, and
  Gupta]{nair2022r3m}
Suraj Nair, Aravind Rajeswaran, Vikash Kumar, Chelsea Finn, and Abhinav Gupta.
\newblock R3m: A universal visual representation for robot manipulation.
\newblock \emph{arXiv preprint arXiv:2203.12601}, 2022.

\bibitem[Shorten and Khoshgoftaar(2019{\natexlab{b}})]{shorten19aug}
Connor Shorten and Taghi~M. Khoshgoftaar.
\newblock A survey on image data augmentation for deep learning.
\newblock \emph{J. Big Data}, 6:\penalty0 60, 2019{\natexlab{b}}.
\newblock \doi{10.1186/s40537-019-0197-0}.
\newblock URL \url{https://doi.org/10.1186/s40537-019-0197-0}.

\bibitem[Kostrikov et~al.(2020)Kostrikov, Yarats, and
  Fergus]{kostrikov2020image}
Ilya Kostrikov, Denis Yarats, and Rob Fergus.
\newblock Image augmentation is all you need: Regularizing deep reinforcement
  learning from pixels.
\newblock \emph{arXiv preprint arXiv:2004.13649}, 2020.

\bibitem[Zafar et~al.(2022)Zafar, Aamir, Mohd~Nawi, Arshad, Riaz, Alruban,
  Dutta, and Almotairi]{zafar2022comparison}
Afia Zafar, Muhammad Aamir, Nazri Mohd~Nawi, Ali Arshad, Saman Riaz,
  Abdulrahman Alruban, Ashit~Kumar Dutta, and Sultan Almotairi.
\newblock A comparison of pooling methods for convolutional neural networks.
\newblock \emph{Applied Sciences}, 12\penalty0 (17):\penalty0 8643, 2022.

\bibitem[Wang et~al.(2022)Wang, Jia, Zhu, Walters, and
  Platt]{wang22equivariant}
Dian Wang, Mingxi Jia, Xupeng Zhu, Robin Walters, and Robert Platt.
\newblock On-robot policy learning with o(2)-equivariant {SAC}.
\newblock \emph{CoRR}, abs/2203.04923, 2022.
\newblock \doi{10.48550/arXiv.2203.04923}.
\newblock URL \url{https://doi.org/10.48550/arXiv.2203.04923}.

\bibitem[Deng et~al.(2021)Deng, Litany, Duan, Poulenard, Tagliasacchi, and
  Guibas]{deng2021vector}
Congyue Deng, Or~Litany, Yueqi Duan, Adrien Poulenard, Andrea Tagliasacchi, and
  Leonidas~J Guibas.
\newblock Vector neurons: A general framework for so (3)-equivariant networks.
\newblock In \emph{Proceedings of the IEEE/CVF International Conference on
  Computer Vision}, pages 12200--12209, 2021.

\bibitem[Kapelyukh et~al.(2022)Kapelyukh, Vosylius, and
  Johns]{kapelyukh2022dall}
Ivan Kapelyukh, Vitalis Vosylius, and Edward Johns.
\newblock Dall-e-bot: Introducing web-scale diffusion models to robotics.
\newblock \emph{arXiv preprint arXiv:2210.02438}, 2022.

\bibitem[Mandi et~al.(2022)Mandi, Bharadhwaj, Moens, Song, Rajeswaran, and
  Kumar]{mandi2022cacti}
Zhao Mandi, Homanga Bharadhwaj, Vincent Moens, Shuran Song, Aravind Rajeswaran,
  and Vikash Kumar.
\newblock Cacti: A framework for scalable multi-task multi-scene visual
  imitation learning.
\newblock \emph{arXiv preprint arXiv:2212.05711}, 2022.

\bibitem[Deitke et~al.(2022)Deitke, VanderBilt, Herrasti, Weihs, Salvador,
  Ehsani, Han, Kolve, Farhadi, Kembhavi, and Mottaghi]{deitke22procthor}
Matt Deitke, Eli VanderBilt, Alvaro Herrasti, Luca Weihs, Jordi Salvador, Kiana
  Ehsani, Winson Han, Eric Kolve, Ali Farhadi, Aniruddha Kembhavi, and Roozbeh
  Mottaghi.
\newblock Procthor: Large-scale embodied {AI} using procedural generation.
\newblock \emph{CoRR}, abs/2206.06994, 2022.
\newblock \doi{10.48550/arXiv.2206.06994}.
\newblock URL \url{https://doi.org/10.48550/arXiv.2206.06994}.

\bibitem[Tobin et~al.(2017)Tobin, Fong, Ray, Schneider, Zaremba, and
  Abbeel]{tobin2017domain}
Josh Tobin, Rachel Fong, Alex Ray, Jonas Schneider, Wojciech Zaremba, and
  Pieter Abbeel.
\newblock Domain randomization for transferring deep neural networks from
  simulation to the real world.
\newblock In \emph{2017 IEEE/RSJ international conference on intelligent robots
  and systems (IROS)}, pages 23--30. IEEE, 2017.

\bibitem[Tremblay et~al.(2018)Tremblay, Prakash, Acuna, Brophy, Jampani, Anil,
  To, Cameracci, Boochoon, and Birchfield]{tremblay2018training}
Jonathan Tremblay, Aayush Prakash, David Acuna, Mark Brophy, Varun Jampani, Cem
  Anil, Thang To, Eric Cameracci, Shaad Boochoon, and Stan Birchfield.
\newblock Training deep networks with synthetic data: Bridging the reality gap
  by domain randomization.
\newblock In \emph{Proceedings of the IEEE conference on computer vision and
  pattern recognition workshops}, pages 969--977, 2018.

\bibitem[Tobin et~al.(2018)Tobin, Biewald, Duan, Andrychowicz, Handa, Kumar,
  McGrew, Ray, Schneider, Welinder, et~al.]{tobin2018domain}
Josh Tobin, Lukas Biewald, Rocky Duan, Marcin Andrychowicz, Ankur Handa, Vikash
  Kumar, Bob McGrew, Alex Ray, Jonas Schneider, Peter Welinder, et~al.
\newblock Domain randomization and generative models for robotic grasping.
\newblock In \emph{2018 IEEE/RSJ International Conference on Intelligent Robots
  and Systems (IROS)}, pages 3482--3489. IEEE, 2018.

\bibitem[Akkaya et~al.(2019)Akkaya, Andrychowicz, Chociej, Litwin, McGrew,
  Petron, Paino, Plappert, Powell, Ribas, et~al.]{akkaya2019solving}
Ilge Akkaya, Marcin Andrychowicz, Maciek Chociej, Mateusz Litwin, Bob McGrew,
  Arthur Petron, Alex Paino, Matthias Plappert, Glenn Powell, Raphael Ribas,
  et~al.
\newblock Solving rubik's cube with a robot hand.
\newblock \emph{arXiv preprint arXiv:1910.07113}, 2019.

\bibitem[Handa et~al.(2022)Handa, Allshire, Makoviychuk, Petrenko, Singh, Liu,
  Makoviichuk, Van~Wyk, Zhurkevich, Sundaralingam, et~al.]{handa2022dextreme}
Ankur Handa, Arthur Allshire, Viktor Makoviychuk, Aleksei Petrenko, Ritvik
  Singh, Jingzhou Liu, Denys Makoviichuk, Karl Van~Wyk, Alexander Zhurkevich,
  Balakumar Sundaralingam, et~al.
\newblock Dextreme: Transfer of agile in-hand manipulation from simulation to
  reality.
\newblock \emph{arXiv preprint arXiv:2210.13702}, 2022.

\end{thebibliography}
\newpage
\appendices
\section{Real-World Experiments}
\subsection{Real-World Tasks}

We collect 10 pick-and-place tasks in the real world in one single environment, as shown in Figure \ref{fig:robot_task}. All tasks are collected with only 10 demonstrations in one single environment. For each demonstration, we randomly place objects within the work zone of the table. To augment the demonstration, we apply GenAug 100 times per demonstration, resulting in 1000 augmented demonstrations for each task. 

% \subsection{Augmentation meshes}
\subsection{Baselines in the real-world}
\begin{figure}[!h]
    \centering
    \includegraphics[width=0.48\textwidth]{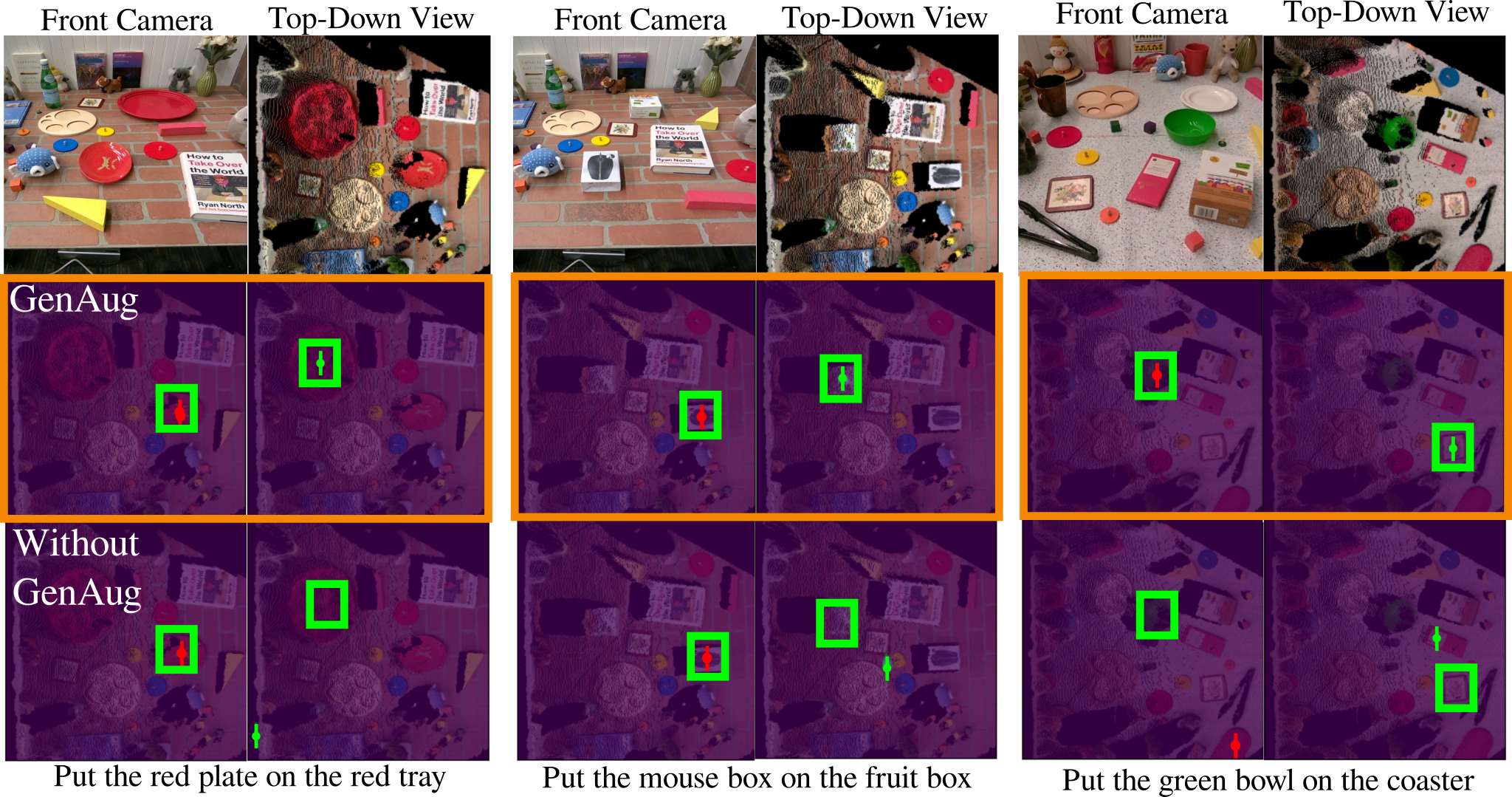}
    \vspace{-0.5em}\caption{Comparison between training with and without GenAug by comparing pick and place affordances predicted by two models. GenAug significantly improves generalization over unseen environments and objects compared to training without GenAug. pick affordances are highlighted in red, and place affordances are highlighted in green. Ground truth locations are represented in green boxes}
    \vspace{-0.5em}
    \label{fig:baseline_realworld}
\end{figure}

We use the ground truth of pick and place locations to evaluate the performance of affordance prediction trained with GenAug and without, as shown in Figure \ref{fig:baseline_realworld}

\subsection{Failure modes}
\begin{figure}[!h]
    \centering
    \includegraphics[width=0.46\textwidth]{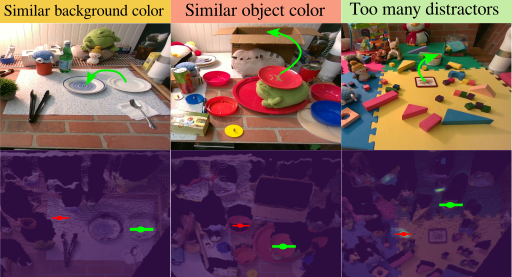}
    \vspace{-0.5em}\caption{Failure cases observed in the real-world setting}
    \vspace{-0.5em}
    \label{fig:failure_mode}
\end{figure}

We observe failure cases usually occur when the background color is similar to the pick or place object. Or one of a few distractors has a very bright color or similar color. We expect this can be improved by increasing the number of augmentations in the training set, such that the training data can have higher coverage of possible combinations of the scenes. 

\subsection{depth-guided diffusion model vs inpainting}
\begin{figure}[!h]
    \centering
    \includegraphics[width=0.45\textwidth]{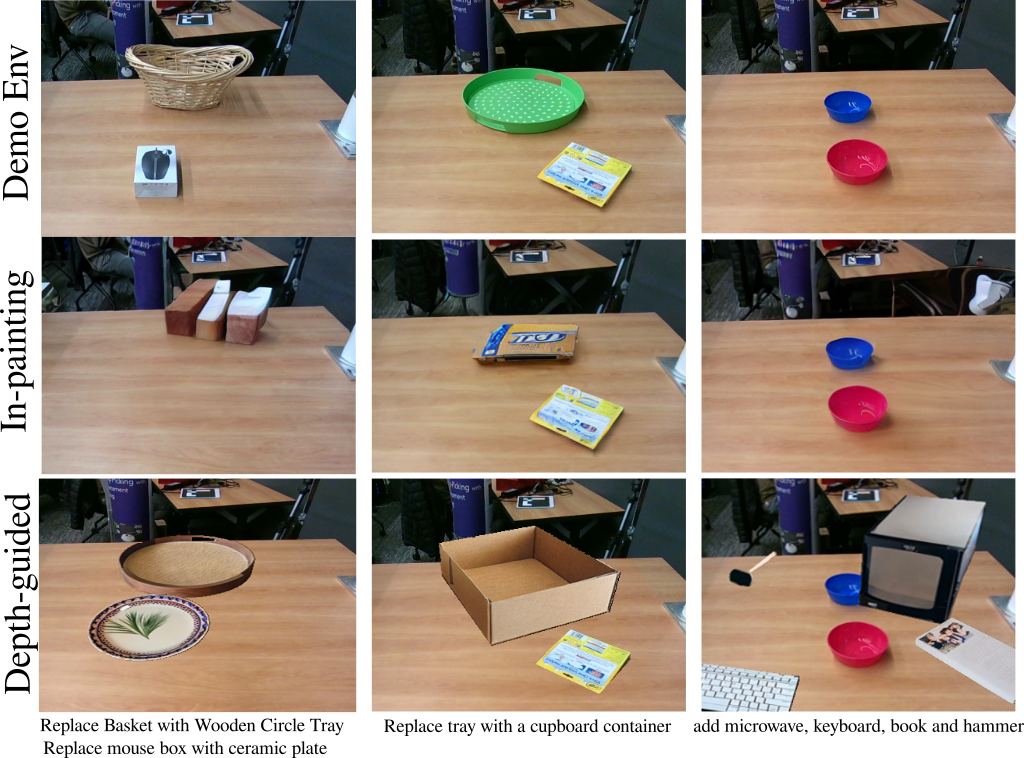}
    \vspace{-0.5em}\caption{Comparison between depth-guided diffusion model with access to predefined 3D meshes and inpainting models.}
    \vspace{-0.5em}
    \label{fig:inpainting}
\end{figure}

We further justify the choice of using a depth-guided diffusion model, as shown in Figure \ref{fig:inpainting}. Directly using the inpainting model often does not result in reasonable visual augmentation. Instead, GenAug uses a depth-guided diffusion model together with predefined 3D meshes, resulting in realistic new objects and scenes.

\subsection{Real-World Unseen Environments}
\begin{figure*}[!h]
    \centering
    \includegraphics[width=\textwidth]{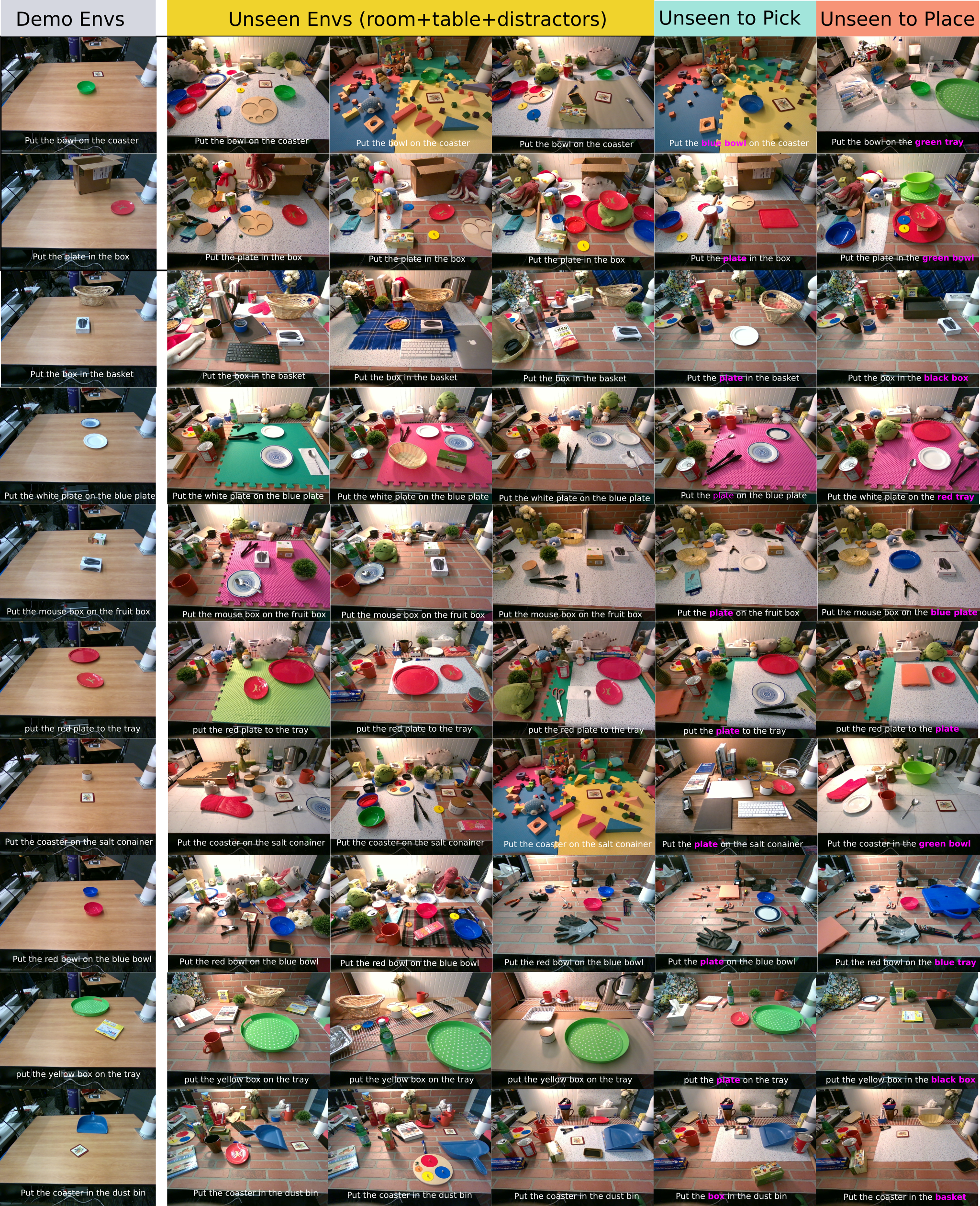}
    \vspace{-0.5em}\caption{Unseen test set in the real-world experiments.}
    \vspace{-0.5em}
    \label{fig:all_robot}
\end{figure*}

In this section, we visualize examples of unseen test scenes that are used for evaluation for all 10 tasks, as shown in Figure \ref{fig:all_robot}

\begin{figure*}[!h]
    \centering
    \includegraphics[width=\textwidth]{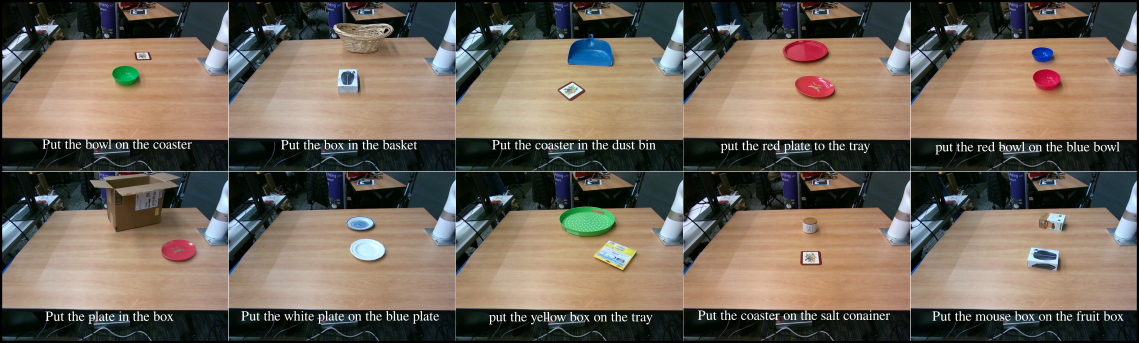}
    \vspace{-0.5em}\caption{Tasks used in the real-world experiments.}
    \vspace{-0.5em}
    \label{fig:robot_task}
\end{figure*}

\subsection{Real-World Evaluation}
We visualize the demonstration environments collected in the real world as well as their corresponding unseen test environments and objects. Please see our website for better visualization \href{https://genaug.github.io}{\color{blue}{https://genaug.github.io}}.

We visualize pick and place affordances predicted by CLIPort trained with GenAug in Figure \ref{fig:robot_experiment}

\section{Simulation Experiments}
\subsection{Tasks}
We perform a large-scale evaluation in simulation. In particular, we collect 1, 10, 100 demonstrations for 5 tasks: "Pack the brown round plate", "Pack the straw hat", "Pack the green and white striped towel", "Pack the grey soccer shoe" and "Pack the porcelain cup", and report the average success rate across all tasks. Following evaluation metrics defined in CLIPort \cite{shridhar2021cliport} and TransporterNet \cite{zeng2020transporter}, the success rate is defined as the total volume of the pick object inside the place object, divided by the total volume of the pick object in the scene.

\subsection{Augmented Dataset in Simulation}
\begin{figure}[!h]
    \centering
    \includegraphics[width=0.49\textwidth]{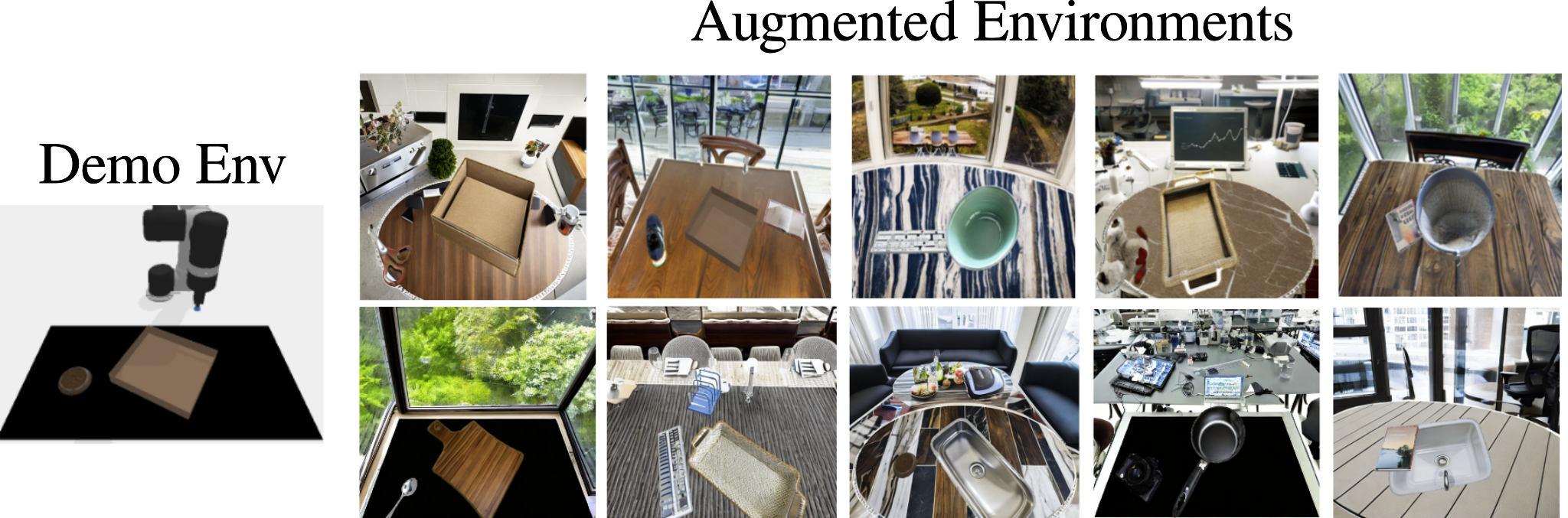}
    \vspace{-0.5em}\caption{Augmented dataset for demonstrations collected in simulation.}
    \vspace{-0.5em}
    \label{fig:sim_aug}
\end{figure}

Given demonstrations from a task collected in simulation, we apply GenAug 100 times for each demonstration. We visualize examples of the augmented dataset in Figure \ref{fig:sim_aug}. 

\begin{figure}[!h]
    \centering
    \includegraphics[width=0.49\textwidth]{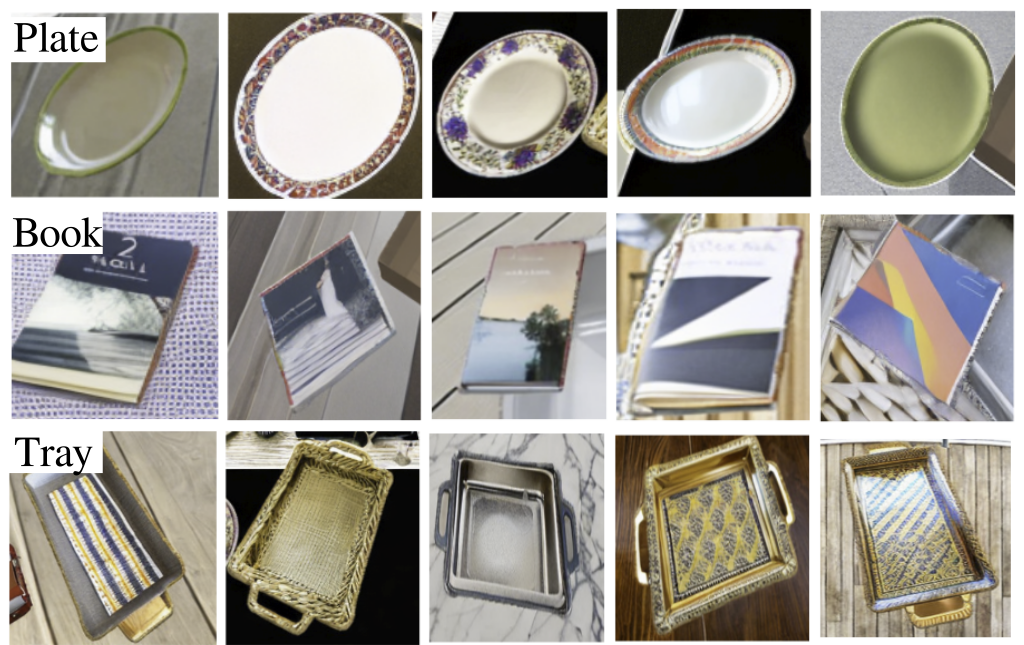}
    \vspace{-0.5em}\caption{Diversity of the appearance of the generated objects}
    \vspace{-0.5em}
    \label{fig:diverse_aug}
\end{figure}

We also observe diverse visual augmentation on the same object template, as shown in Figure \ref{fig:diverse_aug}. Given different text prompts, GenAug is able to generate different and realistic textures.

% \subsection{Details of baselines}
\section{Visualization of Baseline Data augmentation}
\begin{figure}[!h]
    \centering
    \includegraphics[width=0.49\textwidth]{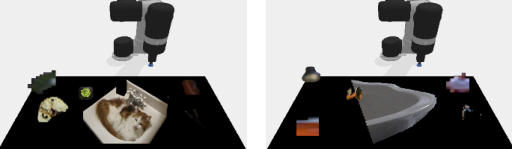}
    \vspace{-0.5em}\caption{Examples of random copy and paste baseline. We extracted queried segmented images from LVIS dataset and paste them directly on the original demonstration image. This usually leads to low-quality and incomplete image generation.}
    \vspace{-0.5em}
    \label{fig:random_paste}
\end{figure}

We visualize some examples of randomly copying and pasting segmented images from LVIS dataset \cite{gupta2019lvis}, as shown in Figure \ref{fig:random_paste}.

We observe this baseline often results in unrealistic, low-quality image generation, which is not usually matching observations during test time in both real-world and simulation. 

\section{Visualizations for Real-World experiments}
\begin{figure*}[!h]
    \centering
    \includegraphics[width=\textwidth]{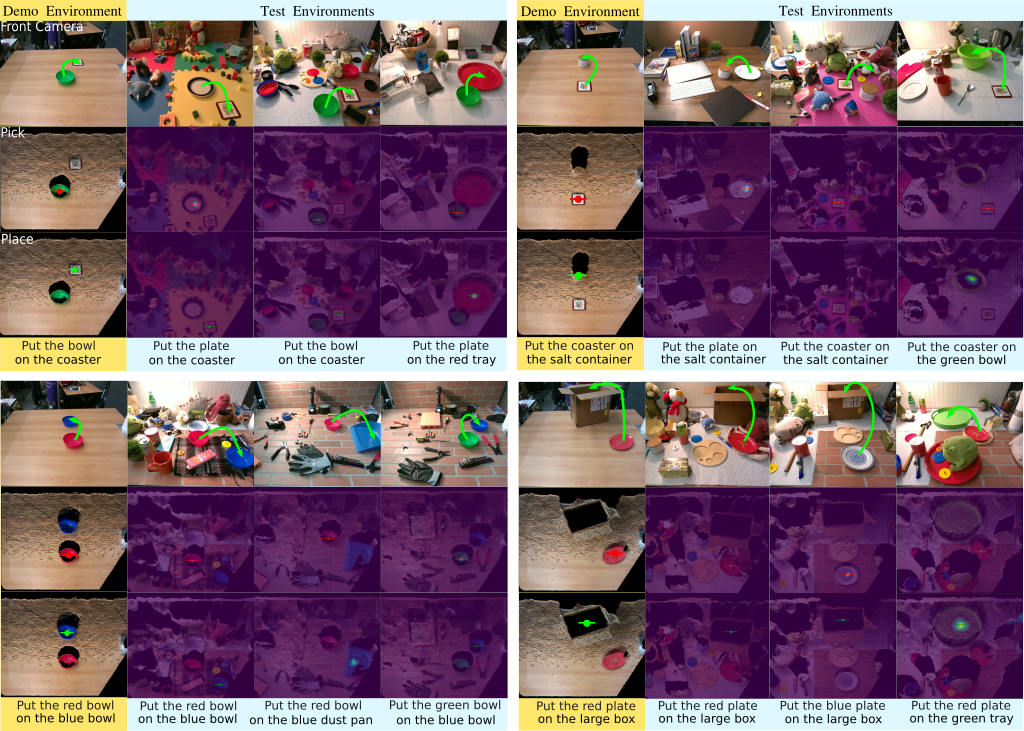}
    \vspace{-0.5em}\caption{Prediction of pick and place locations on various tasks with GenAug}
    \vspace{-0.5em}
    \label{fig:robot_experiment}
\end{figure*}

\begin{figure}[!h]
    \centering
    \includegraphics[width=0.49\textwidth]{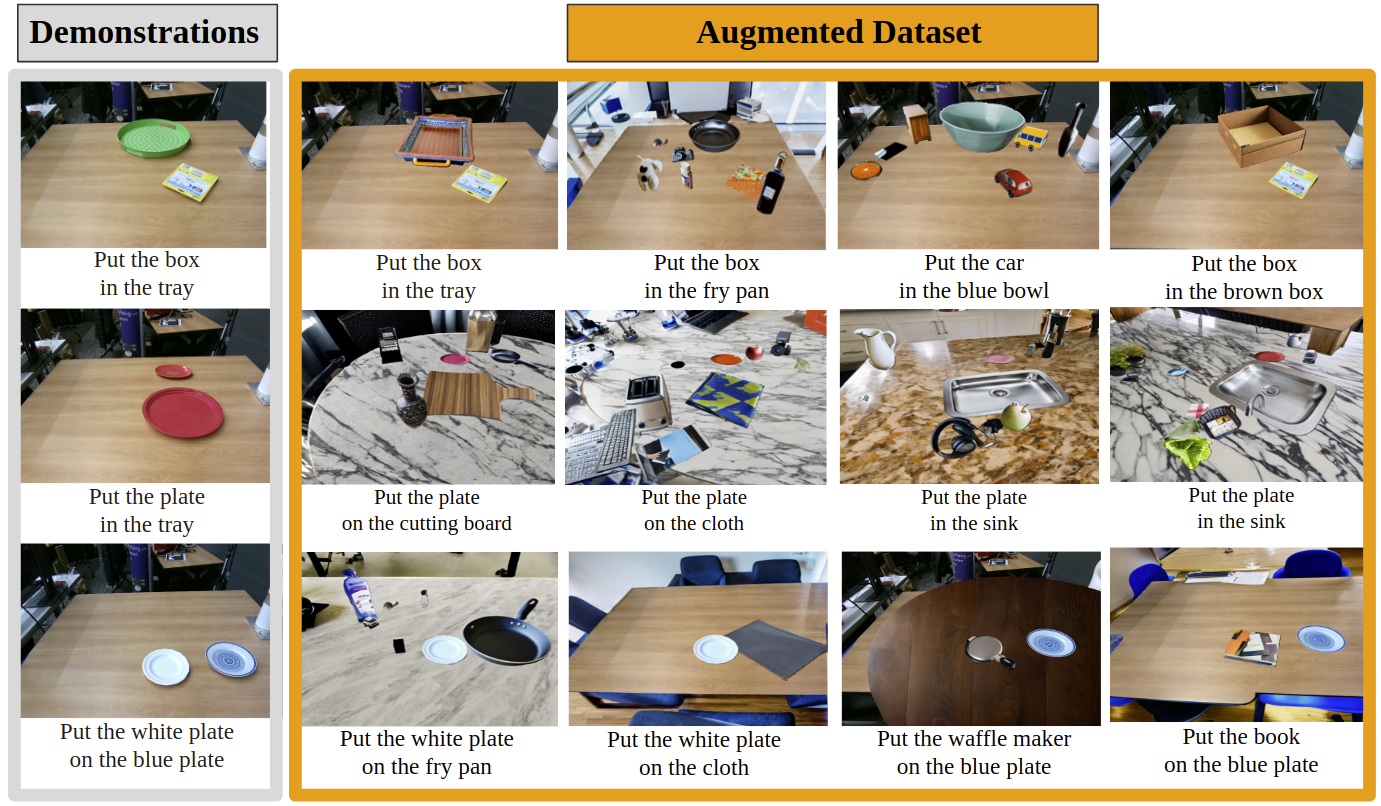}
    \vspace{-0.5em}\caption{Examples of augmented dataset given observations of demonstrations collected in a simple environment.}
    \vspace{-0.5em}
    \label{fig:augmented_dataset}
\end{figure}

\begin{figure}[!h]
    \centering
    \includegraphics[width=0.49\textwidth]{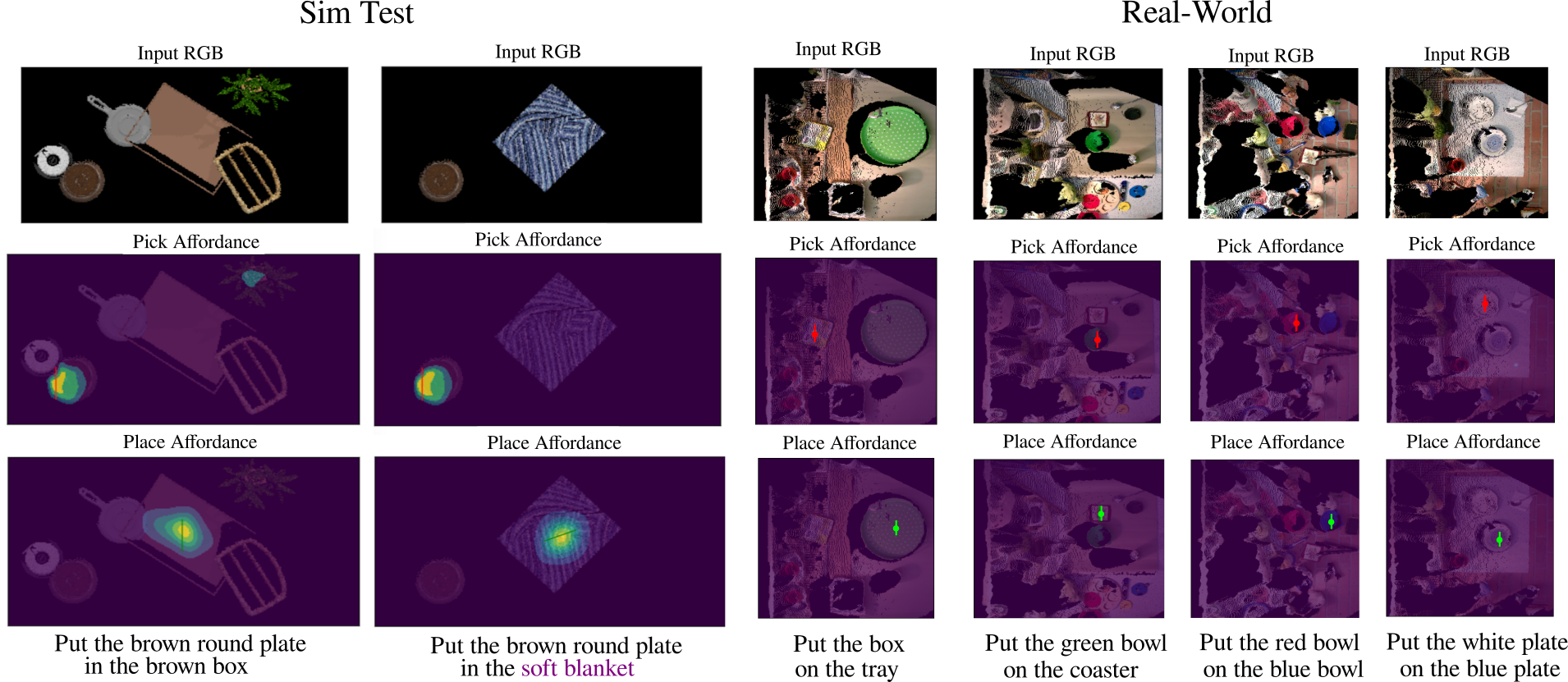}
    \vspace{-0.5em}\caption{Pick and Place affordance predicted by CLIPort that trained on GenAug on unseen environments and objects in simulation and the real world.}
    \vspace{-0.5em}
    \label{fig:affordance}
\end{figure}

In this section, we visualize more examples of applying GenAug on demonstrations collected in the real world, as shown in Figure \ref{fig:augmented_dataset}. We train CLIPort \cite{shridhar2021cliport} with such a dataset and evaluate unseen environments and objects for 10 tasks. We further show affordance predictions in Figure \ref{fig:robot_experiment} and Figure \ref{fig:affordance}.

\end{document}